\documentclass[lettersize,journal]{IEEEtran}
\usepackage{amsmath,amsfonts}
\usepackage{algorithmic}
\usepackage{algorithm}
\usepackage{array}
\usepackage[caption=false,font=normalsize,labelfont=sf,textfont=sf]{subfig}
\usepackage{textcomp}
\usepackage{stfloats}
\usepackage{url}
\usepackage{verbatim}
\usepackage{graphicx}
\usepackage{cite}
\usepackage{makecell}
\usepackage{fancyhdr}
\usepackage{amssymb}
\usepackage{enumitem}
\begin{document}

\title{Deep Learning for Anomaly Detection in Log Data: A Survey}

\author{Max Landauer, Sebastian Onder, Florian Skopik, and 
	Markus Wurzenberger 
	\IEEEcompsocitemizethanks{ Manuscript published in the Machine Learning with Applications, vol. 12 (2023) under the CC BY license. \\ \IEEEcompsocthanksitem https://doi.org/10.1016/j.mlwa.2023.100470 \\ M. Landauer, S. Onder, F. Skopik, and M. Wurzenberger are with the Center for Digital Safety and Security, AIT Austrian Institute of Technology, Vienna, Austria.\protect\\
		E-mail: firstname.lastname@ait.ac.at
		}
}



\maketitle

\begin{abstract}
Automatic log file analysis enables early detection of relevant incidents such as system failures. In particular, self-learning anomaly detection techniques capture patterns in log data and subsequently report unexpected log event occurrences to system operators without the need to provide or manually model anomalous scenarios in advance. Recently, an increasing number of approaches leveraging deep learning neural networks for this purpose have been presented. These approaches have demonstrated superior detection performance in comparison to conventional machine learning techniques and simultaneously resolve issues with unstable data formats. However, there exist many different architectures for deep learning and it is non-trivial to encode raw and unstructured log data to be analyzed by neural networks. We therefore carry out a systematic literature review that provides an overview of deployed models, data pre-processing mechanisms, anomaly detection techniques, and evaluations. The survey does not quantitatively compare existing approaches but instead aims to help readers understand relevant aspects of different model architectures and emphasizes open issues for future work.
\end{abstract}


\section{Introduction} \label{intro}

\IEEEPARstart{L}{og} files provide a rich source of information when it comes to monitoring computer systems. Thereby, the majority of log events are usually generated as consequences of normal system operations, such as starting and stopping of processes, restarting of virtual machines, users accessing resources, etc. However, applications also produce logs when faulty or otherwise undesired system states occur, for example, failed processes, availability issues, or security incidents. These traces of unexpected and possibly unsafe system activities are important for system operators that timely need to act upon them to prevent or diminish system damage and avoid adverse cascading effects.

The main problem for this kind of log file analysis is that it is non-trivial to identify these relevant log events within the much larger number of less interesting traces of standard system usage. In particular, the sheer amount of logs produced by modern applications renders manual analysis infeasible and necessitates automatic mechanisms \cite{wang2021log}. Unfortunately, manually coded signatures and rules that search for specific keywords in logs only have limited applicability and are not suitable for scenarios that are not known beforehand \cite{liao2013intrusion}. It is therefore necessary to deploy anomaly detection techniques that automatically learn models representing the normal baseline of system behavior and subsequently disclose any deviations from these models as possibly adverse activities that require attention by human operators.

Machine learning provides many viable techniques for the purpose of anomaly detection in log files and many different approaches have been proposed in the past, including clustering \cite{landauer2020system} and workflow mining \cite{he2016experience}, statistical analysis of event parameters \cite{kruegel2003anomaly}, time-series analysis to recognize changes of event frequencies \cite{landauer2018dynamic}, and many more \cite{chandola2009anomaly, liao2013intrusion}. Recently, researchers started using deep neural networks for log-based anomaly detection in an attempt to repeat the successes of deep learning from image and speech recognition that outperform conventional machine learning methods \cite{lecun2015deep}. However, as system log events are generally unstructured and involve intricate dependencies, it is non-trivial to prepare the data in a way to enable ingestion by neural networks and extract features that are relevant for detection. Moreover, the wide variety of existing deep learning architectures such as recurrent or convolutional neural networks makes it difficult to select an appropriate model for a specific use-case at hand and understand their respective requirements on the format and properties of the input data.

To the best of our knowledge there is currently only a limited overview of the state-of-the-art of log-based anomaly detection with deep learning \cite{chen2021experience, le2022log, yadav2020survey, kwon2019survey}. As a consequence it is difficult to understand what features are suitable to be extracted from raw log data, how these features could be transformed into a format that is adequate to be ingested by neural networks, and which model architectures are appropriate for detecting specific patterns in logs. Existing surveys only compare few anomaly detection approaches and focus mainly on sequential patterns in log data \cite{chen2021experience, le2022log, yadav2020survey}, present broad studies on system log data analysis that do not sufficiently cover deep learning models and challenges \cite{bhanage2021infrastructure, zhao2021empirical}, or focus on network traffic rather than system log data \cite{kwon2019survey}.

We therefore carry out a systematic literature review on deep learning for anomaly detection in log data. Our main focus is thereby to survey scientific publications on the deployed model architectures, their respective requirements and transformations for handling unstructured input log data, the methods used to differentiate between normal and anomalous data samples, and the presented evaluations. The results of this study are beneficial for researchers and industries alike, because a better understanding of challenges and features of different deep learning algorithms avoids pitfalls when developing anomaly detection techniques and eases selection of existing detection systems for both academic and real-world use-cases. Moreover, a detailed investigation of pre-processing strategies is essential to utilize all information available in the logs when carrying out anomaly detection and to understand the influence of data representations on the detection capabilities, in particular, what types of anomalies can be detected under which circumstances. Regarding scientific evaluations, we particularly pay attention to relevant aspects of experiment design, including data sets, metrics, and reproducibility, to point out deficiencies in prevalent evaluation strategies and suggest remedies. Finally, our study also aims to create a work of reference and establish a starting point for future research. We point out that this survey does not quantitatively compare detection performances of the reviewed approaches as only few open-source implementations are available and comparisons of these approaches are already presented in other surveys \cite{chen2021experience, le2022log}. In alignment with the aforementioned goals we formulate the research questions of this survey as follows: 

\begin{itemize}[leftmargin=11mm]
	\item[RQ1:] What are the main challenges of log-based anomaly detection with deep learning?
	\item[RQ2:] What state-of-the-art deep learning algorithms are typically applied?
	\item[RQ3:] How is log data pre-processed to be ingested by deep learning models?
	\item[RQ4:] What types of anomalies are detected and how are they identified as such?
	\item[RQ5:] How are the proposed models evaluated?
	\item[RQ6:] To what extent do the approaches rely on labeled data and support incremental learning?
	\item[RQ7:] To what extent are the presented results reproducible in terms of availability of source code and used data?
\end{itemize}

The remainder of this paper is structured as follows. Section \ref{background} first explains the terms deep learning, log data, and anomaly detection, and then provides an overview of common challenges. We explain our methodology for selecting relevant publications and carrying out the survey in Sect. \ref{method}. Section \ref{results} presents all results of our survey in detail. We discuss these results and answer our research questions in Sect. \ref{discussion}. Finally, Sect. \ref{conclusion} concludes this paper.

\section{Background} \label{background}

In this section we first clarify some general concepts and terms relevant for anomaly detection in log data based on deep learning. We then outline scientific challenges that are specific to that research field.

\subsection{Preliminary Definitions}

The study carried out in this paper hinges on an understanding of three main concepts: deep learning, log data, and anomaly detection. However, the exact characteristics and consequential requirements of the respective fields may be used differently across research areas and existing literature. In the following, we therefore describe the basic properties of these three concepts.

\subsubsection{Deep Learning}

Artificial neural networks (ANN) have been developed in an attempt to recreate biological information processing systems in the form of connected communication nodes. For this purpose, varying numbers of nodes are arranged in sequences of layers, in particular, an input layer that reads in the data, several hidden layers connecting neighboring layers with weighted edges, and an output layer. Nodes activate when receiving specific signals on their connected edges, which in turn generates the input for subsequent layers. The main idea is that such networks are capable of recognizing non-linear structures in the input data and subsequently classifying the processed instances through training, which involves minimizing the error of classifications by adjusting the weights of edges accordingly. Thereby, ANN enable supervised training where labels for all classes are available (i.e., data samples are marked with labels such as normal or anomalous), semi-supervised training where labels of some classes are available, as well as unsupervised learning where no labels are available. 

In general, deep learning algorithms are understood as neural networks with multiple hidden layers. Several different architectures of deep neural networks have been proposed in the past, such as recurrent neural networks (RNN) for sequential input data. Deep learning has been shown to outperform conventional machine learning methods (e.g., support vector machines or decision trees) and even human experts in many application areas such as image classification, speech recognition, and many more \cite{lecun2015deep, sarker2021deeplearning}.

\subsubsection{Log Data} \label{logdatadef}

Log data are a chronological sequence of single- or multi-line events generated by applications to permanently capture specific system states, in particular, for manual forensic analysis in case that failures or other unexpected incidents occur. Log events are usually available in textual form and range from structured vectors (e.g., comma-separated values) over semi-structured objects (e.g., key-value pairs) to unstructured human-readable messages with heterogeneous event types. Despite the fact that no unified log format exists, log events usually contain their generation time stamp as one of their event parameters. Other parameters that are sometimes present in different types of log data are logging levels (e.g., \textit{INFO} or \textit{ERROR}) or process identifiers that link sequences of related events \cite{landauer2020system}.

While single log events describe (part of) the system state in one particular point in time, groups of log events represent the dynamic workflows of the underlying program logic. The reason for this is that log events are generated by print statements purposefully placed by software developers throughout their code to support understanding of program activities and debugging. These statements comprise of static parts, i.e., hard-coded strings, and variable parts, i.e., parameters that are dynamically determined during program runtime. In the past, a large amount of research was directed towards automatic extraction of so-called log keys (also known as log signatures, log templates, or simply log events) that represent templates for the original print statements and enable parsing of logs \cite{zhu2019tools}. These parsers allow to derive values from logs that are more suitable to be used for subsequent analysis than unstructured log messages, in particular, through (i) assignment of event type identifiers to log events and (ii) extraction of parameters from log messages \cite{bao2018execution}. 

\subsubsection{Anomaly Detection} \label{anomalydetection}

Anomalies are those instances in a data set that exhibit rare or otherwise unexpected characteristics and thus stand out from the rest of the data \cite{chandola2009anomaly}. For the purpose of detection, the conformity of these data instances is usually measured through one or more continuous or categorical attributes that are associated with each instance and enable the computation of similarity metrics. For independent data, it is sufficient to declare single or small groups of instances with high dissimilarities to all other data points as outliers, which are also referred to as point anomalies. For all data where instances are not independent from each other, e.g., all kinds of ordered data including log data, two additional types of anomalies occur. First, contextual anomalies are instances that are only anomalous with respect to the context they occur (but not otherwise), such as the time of occurrence. For example, consider the start of a daily executed data backup procedure that suddenly takes place outside of the scheduled times. Second, collective anomalies are groups of instances that are only anomalous due to their combined occurrence (but not individually), such as a specific sequence of log events. 

An implicit assumption of most anomaly detection techniques is that the analyzed data holds far fewer anomalies than normal instances. This enables that detection takes place in a fully unsupervised manner, i.e., no labeled data is necessary to train the models. However, many scientific approaches instead pursue semi-supervised detection, where training data containing only normal instances are available and evaluation then takes place on a test data set comprising both normal and anomalous instances. The main advantages of semi-supervised operation is that anomalous instances are not learned by the models and that it is often relatively simple to gather normal data, while modeling and labeling anomalies is less straightforward. Accordingly, approaches for supervised anomaly detection have lower applicability and are comparatively rare \cite{he2016experience}.

\subsection{Challenges} \label{challenges}

Log-based anomaly detection has been an active field of research for decades. Thereby, most of the presented approaches rely on conventional machine learning techniques. However, the last few years have seen a strong increase of approaches that leverage deep learning to disclose anomalous log events that relate to unexpected system behavior. In the following, we summarize the main challenges that need to be overcome for effective and applicable detection.

\begin{itemize}
	\item \textbf{Data representation}. Deep learning systems generally consume structured and numeric input data. It is non-trivial to feed log data into neural networks as they frequently involve a mix of heterogeneous event types, unstructured messages, and categorical parameters \cite{yadav2020survey, chalapathy2019deep}. 
	\item \textbf{Data instability}. As applications evolve, new log event types may occur that differ from the ones in the training data. In addition, the observed system behavior patterns are subject to change as technological environments and their utilization vary over time. Deep learning systems therefore need to incrementally update their models and adapt their baseline for normal system behavior to enable real-time detection \cite{zhang2019robust, yadav2020survey, chalapathy2019deep, chen2021experience, le2022log}.
	\item \textbf{Class imbalance}. Anomaly detection inherently assumes that normal events outnumber anomalous ones. Many approaches based on neural networks are known to perform sub-optimally for imbalanced data sets \cite{chalapathy2019deep, le2022log}.
	\item \textbf{Anomalous artifact diversity}. Manifestations of anomalies affect log events as well as parameters thereof in various ways, including changes of sequential patterns, frequencies, correlations, inter-arrival times, etc. Detection techniques are often designed only for properties of specific anomaly types and are therefore not generally applicable.
	\item \textbf{Label availability}. As anomalies represent unexpected system behavior, there are generally no labeled anomaly instances available for training. This restricts applications to semi- and unsupervised deep learning systems, which are known to achieve lower detection performance than supervised approaches \cite{chen2021experience, le2022log}.
	\item \textbf{Stream processing}. Logs are generated as a continuous stream of data. To enable on-the-fly monitoring rather than forensic analysis, deep learning systems need to be designed for single-pass data processing when it comes to detection and model updating \cite{landauer2020system, le2022log}.
	\item \textbf{Data volume}. Log data is generated in high volumes, with some systems producing millions \cite{xu2009detecting} or even billions \cite{mi2013toward} of events daily. Efficient algorithms are required to ensure real-time processing in practical applications, in particular, when running on machines with few computational resources such as edge devices.
	\item \textbf{Interleaving logs}. Sequences of related log events may be interleaving each other when many processes operate simultaneously or distributed logs are collected centrally. It is non-trivial to retrieve the original event sequences when events lack session identifiers \cite{le2022log}.
	\item \textbf{Data quality}. Low data quality may be the result of improper log collection or technical issues during log generation and cause negative effects on machine learning effectiveness. Common problems involve incorrect time stamp information, event ordering, missing events, duplicated records, mislabeled events, etc. \cite{suriadi2017event, fischer2020enhancing}.
	\item \textbf{Model explainability}. Approaches based on neural networks generally suffer from a lower explainability than conventional machine learning methods. Difficulties to understand the reasons behind both correct and incorrect classifications are especially problematic when it comes to making justified decisions in response to critical system behavior or security incidents \cite{chalapathy2019deep, chen2021experience}.
\end{itemize}

\section{Survey Method} \label{method}

This section outlines the method that was used to carry out the systematic literature review. We first describe our strategy for collecting relevant literature and then present the evaluation criteria that we used to analyze the retrieved papers.

\subsection{Search Strategy}

In this section we describe the process of gathering relevant publications to be included in the survey. First, an initial collection of literature is collected using a web search. Subsequently, relevant papers are selected using inclusion, exclusion, and quality criteria.

\subsubsection{Initial Literature Collection with Search String}

\begin{figure}
	\centering
	\includegraphics[width=1\columnwidth]{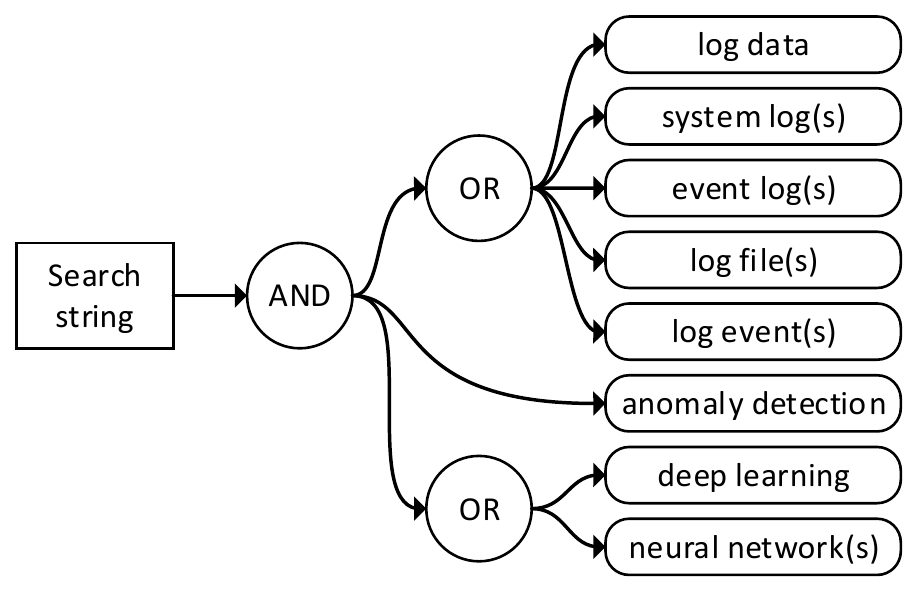}
	\caption{Composition of the search string used to retrieve relevant literature.}
	\label{fig:searchstring}
\end{figure}

In order to obtain an initial set of approaches from the state-of-the-art, we assemble a search string to query common databases for scientific publications. In particular, we design the search string so that only publications containing the three main concepts relevant for this survey are retrieved: log data, anomaly detection, and deep learning. Since some publications use different terminology and to decrease the likelihood that relevant publications are missed, we also use the terms ``system log(s)'', ``event log(s)'', ``log file(s)'', ``log event(s)'' as alternatives for ``log data'', and ``neural network(s)'' as an alternative for ``deep learning''. We omit the term ``log'' without any other word as it yields many results that contain logarithms but are not relevant for our study. Figure \ref{fig:searchstring} displays the final search string.

We then use this search string to gather publications from the following databases: Science Direct\footnote{\url{https://www.sciencedirect.com/}}, Scopus\footnote{\url{https://www.scopus.com/}}, SpringerLink\footnote{\url{https://link.springer.com/}}, ACM Digital Library\footnote{\url{https://dl.acm.org/}}, IEEE Xplore\footnote{\url{https://ieeexplore.ieee.org/}}, Google Scholar\footnote{\url{https://scholar.google.com/}}, and Web of Science\footnote{\url{https://www.webofscience.com/}}. The search was conducted in January of 2022 and returned a total of 2925 publications. In the following section, we describe our method for selecting relevant publications from this set.

\subsubsection{Selection of Relevant Publications}

To sort out publications that are not relevant for this survey and reduce the set of publications to a manageable size, we define multiple selection criteria and apply them on our initial collection. Our main criterion for including the publication in the survey is as follows: \textit{The model proposed in the publication applies deep learning techniques (i.e., a multi-layered neural network) for anomaly detection in heterogeneous and unstructured log data}. Moreover, we define several exclusion criteria that we use to omit publications with low relevance or otherwise inappropriate format. The list of exclusion criteria is as follows.

\begin{itemize}
	\item There is no indication stated in the paper that the presented approach is applicable or designed for application with log data. Our survey does not attempt to adopt methods from other domains for log data analysis.
	\item There exists a more recent publication that presents the same or a similar study.	The purpose of this criteria is to ensure that the most up-to-date versions of specific approaches are analyzed.
	\item The publication only applies an existing approach without novel modifications from the original concept.
	\item The publication is in any language other than English.
	\item The publication is not available in electronic form.
	\item The publication is a book, technical report, lecture note, presentation, review, or thesis.
	In these cases, the corresponding conference or journal publications were reviewed if possible.
\end{itemize}

Note that we do not constraint the publications to a specific time range in order to avoid missing any older publications that are nonetheless relevant for this survey. However, we aim to omit publications of generally lower quality that do not meet the minimum scientific standards. We therefore only select publications that meet the following quality criteria in addition to our main inclusion criterion.

\begin{itemize}
	\item The purpose of the study and its findings are explicitly stated, e.g., the design of a deep learning system for the purpose of anomaly detection is stated as the main contribution of the paper.
	\item The applied deep learning models and their parameters are rigorously described, i.e., it is clear to the reader what type of deep learning model was selected and how its layout was designed, e.g., how many layers it comprises.
	\item The publication includes a sound evaluation of the presented approach, comprising a convincing motivation for the design of the conducted experiments, a comprehensive description of the evaluation process and overall setup, an explanation for choosing the captured metrics, and a detailed discussion of the gathered results and their implications. 
	\item The data sets used for evaluating the approach are referenced or described. This criteria ensures that our survey does not include publications presenting potentially misleading findings originating from data sets that are inadequate for anomaly detection. 
	\item Visualizations are clear and readable. Incomprehensible presentation of results are misleading and lack scientific value. 
\end{itemize}

Our selection procedure is a two-stage process. First, we reduce the initial collections of publications using the inclusion and exclusion criteria based on the title and abstract of each paper. After this stage 331 publications remained. In the second stage, we carry out the selection using all aforementioned criteria based on the contents of each paper. We eventually obtained 62 papers that were included in this survey. The following section outlines our method for analyzing these publications.

\subsection{Reviewed features} \label{features}

To ensure that we analyze the selected publications on a common scheme and address our research questions, we formulate a list of features that we assess for each paper. The following set of questions concerns the applied deep learning (DL) model and mode of operation.

\begin{itemize}[leftmargin=11mm]
	\item[DL-1:] Which deep learning models are used?
	\item[DL-2:] Which training loss functions are applied?
	\item[DL-3:] Does the approach support online or incremental\footnote{Online or incremental processing refers to single-pass procedures where the runtime grows approximately linear with the number of processed lines.} learning?
	\item[DL-4:] Does training take place in un-, semi-, or supervised manner?
\end{itemize}

We then analyze the different ways how the reviewed approaches feed raw log data into deep learning models. The following questions therefore address the pre-processing (PP) and transformation of logs into numeric vector or matrix representations.

\begin{itemize}[leftmargin=11mm]
	\item[PP-1:] How are raw logs pre-processed?
	\item[PP-2:] What features are extracted from pre-processed logs? 
	\item[PP-3:] How are extracted features represented as vectors?
\end{itemize}

The next set of questions deals with the anomaly detection (AD). In particular, we are interested in the different types of anomalies to understand whether they are linked to the features extracted from the raw logs and their representations as vectors. 

\begin{itemize}[leftmargin=11mm]
	\item[AD-1:] What types of anomalies are detected by the approach?
	\item[AD-2:] How is the output of the deep neural network\footnote{The output layer of the neural network comprises one or more nodes with certain numeric values.} used for anomaly detection?
	\item[AD-3:] How are anomalies differentiated from normal data samples?
\end{itemize}

Utilizing openly accessible data sets for evaluations as well as publishing source code alongside papers is not only good scientific standard but also essential for others to validate presented results and carry out comparisons. The last set of questions therefore concerns evaluation and reproducibility (ER), in particular, employed evaluation metrics as well as availability of data sets and source code. 

\begin{itemize}[leftmargin=11mm]
	\item[ER-1:] What log data sets are used for evaluating the approach?
	\item[ER-2:] What evaluation metrics are employed?
	\item[ER-3:] Does the evaluation consider runtime performance measurements?
	\item[ER-4:] What approaches are used as benchmarks?
	\item[ER-5:] Are the used data sets publicly available?
	\item[ER-6:] Is the source code of the approach publicly available?
\end{itemize}

All aforementioned questions were assessed for each publication individually. The resulting feature matrix is presented in Table \ref{tab:results} in the following section and serves as the basis for our analyses and discussions.

\section{Survey Results} \label{results}

This section provides the assessments of all reviewed publications with respect to the features outlined in the previous section. We first provide some general information on the meta-data of publications before going over each reviewed feature in detail.

\subsection{Bibliometrics}

This section provides an overview of the distribution of publications per year as well as their citation counts.

\subsubsection{Publications per Year}

\begin{figure}
	\centering
	\includegraphics[width=1\columnwidth]{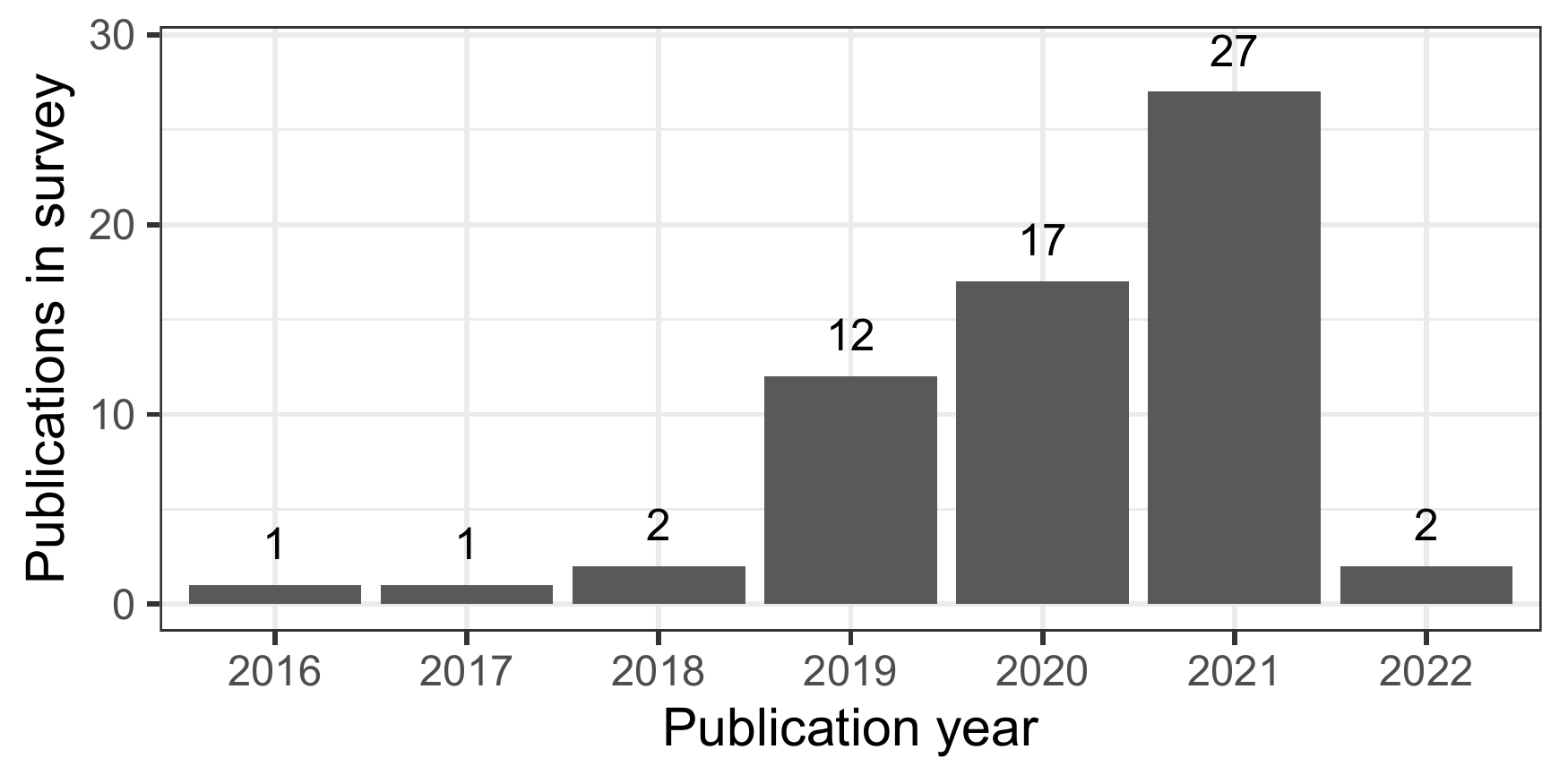}
	\caption{Distribution of the number of publications per year.}
	\label{fig:counts}
\end{figure}

Deep learning for anomaly detection in log data is a relatively new research field that has increasingly gained traction in the last years. Accordingly, a majority of the publications in this research area have only been published in the last two to three years. Figure \ref{fig:counts} shows an overview of the publication years of all publications reviewed for this survey. As expected, 58 out of the 62 considered publications were published in 2019 or later. As the search for relevant literature was carried out in the beginning of 2022, only two publications from that year are included. However, we expect to see an even higher number of publications in 2022 and beyond following the overall trend visible in the plot.

\subsubsection{Citations} \label{citations}

Citation counts are a common  indicator to assess the relevance and influence of publications. We therefore state the top six publication with the highest citation counts (according to Google Scholar) in Table \ref{tab:citations}. As of January 2023, the paper presenting DeepLog by Du et al. \cite{du2017deeplog} that was published in 2017 has the highest citation count and is arguably the most influential of all reviewed publications as they were the first to propose an approach based on deep learning that enables detection of anomalous event sequences in log data. Several of the subsequently published papers rely on the groundwork of DeepLog and it is therefore fair to assume that this paper is at least to some degree responsible for the increase of relevant publications from 2019 and onward that is visible in Fig. \ref{fig:counts}.

Note that the publication by Yang et al. \cite{yang2016log} predates DeepLog \cite{du2017deeplog} but has a significantly lower citation count. The main reason for this is that the paper focuses on the analysis of tokens in single log events, a topic that received far less attention in subsequent research than the analysis of event sequences. This differentiation as well as assessments for all other features stated in Sect. \ref{features} are presented in Table \ref{tab:results}. 

\begin{table*}[]
	\small
	\caption{Top six most cited publications}
	\begin{tabular}{lllll}
		\textbf{Citations} & \textbf{Approach} & \textbf{Year} & \textbf{Authors} & \textbf{Paper Title} \\ \hline \hline
		963 & DeepLog & 2017 & \cite{du2017deeplog} & \makecell[lt]{DeepLog: Anomaly Detection and Diagnosis from System Logs through \\ Deep Learning} \\ \hline
		227 & LogRobust & 2019 & \cite{zhang2019robust} & Robust Log-Based Anomaly Detection on Unstable Log Data \\ \hline
		209 & LogAnomaly & 2019 & \cite{meng2019loganomaly} & \makecell[lt]{LogAnomaly: Unsupervised Detection of Sequential and Quantitative \\ Anomalies in Unstructured Logs} \\ \hline
		92 & - & 2018 & \cite{lu2018detecting} & \makecell[lt]{Detecting Anomaly in Big Data System Logs Using Convolutional Neural \\ Network} \\ \hline
		53 & Logsy & 2020 & \cite{nedelkoski2020self} & Self-Attentive Classification-Based Anomaly Detection in Unstructured Logs \\ \hline
		45 & LogBERT & 2021 & \cite{guo2021logbert} & \makecell[lt]{LogBERT: Log Anomaly Detection via BERT} \\ \hline
	\end{tabular}
	\label{tab:citations}
\end{table*}

\begin{table*}[]
	\renewcommand{\arraystretch}{1.0}
	\setlength{\tabcolsep}{2.0pt}
	\scriptsize
	\caption{Survey results. DL-1: Multi-Layer Perceptron (MLP), Convolutional Neural Network (CNN), Recursive Neural Network (RNN), Autoencoder (AE), Generative Adversarial Network (GAN), Transformer (TF), Attention mechanism (AT), Graph Neural Network (GNN), Evolving Granular Neural Network (EGNN); DL-2: Cross-Entropy (CE), Hyper-Sphere (HS), Mean Squared Error (MSE), Kullback-Leibler Divergence (KL), Marginal Likelihood (ML), Custom Loss Function (CF), Adversarial Training (AT), Not Available (NA); DL-3: Online (ON), Offline (OFF); DL-4: Supervised (SUP), Semi-supervised (SEMI), Unsupervised (UN); PP-1: Log key (KEY), Token (TOK), Combination (COM); PP-2: Token Sequence (TS), Token Count (TC), Event Sequence (ES), Event Count (EC), Parameter (PA), Event Interval Time (EI); PP-3: Event ID sequence (ID), Count Vector (CV), Statistical Feature Vector (FV), Semantic Vector (SV), Positional Embedding (PE), One-Hot Encoding (OH), Embedding Layer/Matrix (EL), Deep Encoded Embedding (DE), Parameter Vector (PV), Time Embedding (TE), Graph (G), Transfer Matrix (TM); AD-1: Outlier (OUT), Sequential (SEQ), Frequency (FREQ), Statistical (STAT); AD-2: Binary Classification (BIN), Input Vector Transformations (TRA), Reconstruction Error (RE), Multi-class Classification (MC), Probability Distribution (PRD), Numeric Vector (VEC); AD-3: Label (LAB), Threshold (THR), Highest Probabilities (TOP).}
	\begin{tabular}{lccccccccccc}
		\textbf{Approach} & \textbf{DL-1} & \textbf{DL-2} & \textbf{DL-3} & \textbf{DL-4} & \textbf{PP-1} & \textbf{PP-2} & \textbf{PP-3} & \textbf{AD-1} & \textbf{AD-2} & \textbf{AD-3} & \textbf{ER-6} \\ \hline \hline
		Baril et al. \cite{baril2020application} (NoTIL)              & RNN & CE & OFF & SEMI &              KEY & ES, EC & CV & FREQ & VEC & THR & NO \\ \hline
		Bursic et al. \cite{bursic2019anomaly}                        & RNN, AE & MSE & OFF & UN &           TOK & PA, TS & DE & OUT & RE & THR & NO \\ \hline
		Catillo et al. \cite{catillo2022autolog} (AutoLog)            & AE & MSE & ON & SEMI &               COM & STAT, TC & FV & FREQ & RE & THR & YES \\ \hline
		Cheansunan et al. \cite{cheansunan2019detecting}              & CNN & NA & OFF & SEMI &              KEY & ES & EL & SEQ & PRD & TOP & NO \\ \hline
		Chen et al. \cite{chenunsupervised}                           & RNN, CNN & NA & OFF & SEMI &         KEY & ES, EC & SV, CV & SEQ, FREQ & VEC, PRD & TOP & NO \\ \hline
		Chen et al. \cite{chen2020logtransfer} (LogTransfer)          & RNN, CNN & NA & OFF & SUP &          KEY & ES & SV & SEQ & BIN & LAB & YES \\ \hline
		Decker et al. \cite{decker2020comparison}                     & EGNN & CF & OFF & SUP &              KEY & EC, STAT & FV & FREQ & MC & THR, LAB & NO \\ \hline
		Du et al. \cite{du2017deeplog} (DeepLog)                      & RNN & CE, MSE & ON & SEMI &          KEY & ES, PA, EI & PV, OH & SEQ & VEC, PRD & THR, TOP & RE \\ \hline
		Du et al. \cite{du2021log} (LogAttention)                     & TF, AM & NA & OFF & SUP &            KEY & ES & SV & SEQ & BIN & LAB & NO \\ \hline
		Farzad et al. \cite{farzad2019log}                            & RNN, AE & CE & OFF & SUP &           TOK & TS & EL & SEQ & BIN & LAB & NO \\ \hline
		Farzad et al. \cite{farzad2020log}                            & RNN, AE, CNN, GAN & CE & OFF & SUP & TOK & TS & EL & OUT & BIN & LAB & NO \\ \hline
		Farzad et al. \cite{farzad2021two}                            & RNN & CE & OFF & SUP &               TOK & TS & EL & OUT & BIN & LAB & NO \\ \hline
		Gu et al. \cite{gu2021research}                               & RNN, AM & CE & OFF & SEMI &          KEY & ES & SV & SEQ & PRD & TOP & NO \\ \hline
		Guo et al. \cite{guo2021anomaly} (FLOGCNN)                    & CNN & CE & OFF & SUP &               COM & TS & SV & SEQ & BIN & LAB & NO \\ \hline
		Guo et al. \cite{guo2021logbert} (LogBERT)                    & TF, AM & CE, HS & OFF & SEMI &           KEY & ES & PE, EL & SEQ & PRD & TOP & YES \\ \hline
		Guo et al. \cite{guo2021translog} (TransLog)                  & TF, AM & CE & OFF & SUP &            KEY & ES & SV & SEQ & BIN & LAB & NO \\ \hline
		Han et al. \cite{han2021unsupervised} (LogTAD)                & RNN, GAN & HS, CF, AT & OFF & UN & KEY & ES & SV & SEQ & TRA & THR & YES \\ \hline
		Hashemi et al. \cite{hashemi2021onelog} (OneLog)              & CNN & CE & OFF & SUP &               TOK & ES & DE & SEQ & BIN & LAB & YES \\ \hline
		Hirakawa et al. \cite{hirakawa2020software}                   & CNN, TF & CF & OFF & UN &            TOK & ES & SV & OUT & BIN & THR & NO \\ \hline
		Huang et al. \cite{huang2020hitanomaly} (HitAnomaly)          & TF, AM & CE & OFF & SUP &            KEY & ES, PA & SV, PV & SEQ & BIN & LAB & NO \\ \hline
		Le et al. \cite{le2021log} (NeuralLog)                        & TF & CE & OFF & SUP &                TOK & ES & SV, PE & SEQ & BIN & LAB & YES \\ \hline
		Li et al. \cite{li2020logspy} (LogSpy)                        & CNN, MLP, AM & CE & OFF & SUP &      KEY & STAT & DE, EL & STAT & BIN & LAB & NO \\ \hline
		Li et al. \cite{li2020swisslog} (SwissLog)                    & RNN, AM & CE & OFF & SUP &           KEY & ES, EI & SV, TE & STAT, SEQ & BIN & LAB & NO \\ \hline
		Liu et al. \cite{liu2021lognads} (LogNADS)                    & RNN & MSE & OFF & SUP &              KEY & ES, PA & SV, PV & SEQ & BIN & LAB & NO \\ \hline
		Lu et al. \cite{lu2018detecting}                              & CNN & NA & OFF & SUP &               KEY & ES & EL & SEQ & BIN & LAB & NO \\ \hline
		Lv et al. \cite{lv2021conanomaly} (ConAnomaly)                & RNN & NA & OFF & SUP &               KEY & ES & SV & SEQ & BIN & THR & NO \\ \hline
		Meng et al. \cite{meng2019loganomaly} (LogAnomaly)            & RNN & NA & OFF & SEMI &              KEY & ES, EC & SV, CV & SEQ, FREQ & VEC, PRD & TOP & RE \\ \hline
		Nedelkoski et al. \cite{nedelkoski2020self} (Logsy)           & TF & HS & OFF & UN &                 TOK & TS & PE, EL & OUT & TRA & THR & RE \\ \hline
		Otomo et al. \cite{otomo2019latent}                           & AE & KL, ML, CF & OFF & SEMI &       KEY & ES, EC & CV, OH & FREQ & TRA & THR & NO \\ \hline
		Ott et al. \cite{ott2021robust}                               & RNN & CE, MSE & OFF & SEMI &         KEY & ES & SV & SEQ & VEC, PRD & THR, TOP & NO \\ \hline
		Patil et al. \cite{patil2019explainable}                      & RNN & CE & OFF & SUP &               KEY & ES & OH & SEQ & BIN & LAB & NO \\ \hline
		Qian et al. \cite{qian2020anomaly} (VeLog)                    & AE & CF & ON & SEMI &                KEY & ES, EC & ID, CV & SEQ, FREQ & RE & THR & NO \\ \hline
		Studiawan et al. \cite{studiawan2021anomaly}                  & AE & MSE & OFF & SEMI &              KEY & EC, STAT, EI, TC & FV & STAT, FREQ & RE & THR & NO \\ \hline
		Studiawan et al. \cite{studiawan2020anomaly} (pylogsentiment) & RNN & CE & OFF & SUP & TOK & TS & SV & OUT & BIN & LAB & YES \\ \hline
		Sun et al. \cite{sun2020context} (AllContext)                 & RNN, AM & CE & OFF & SUP &           KEY & ES & SV & OUT, SEQ & BIN, MC & LAB & NO \\ \hline
		Sundqvist et al. \cite{sundqvist2020boosted} (BoostLog)       & RNN & NA & ON & SEMI &               KEY & ES & ID & SEQ & PRD & THR & NO \\ \hline
		Syngal et al. \cite{syngal2021server}                         & RNN, AE & CE & OFF & SUP &           TOK & ES, EC & CV, OH & SEQ, FREQ & RE & THR & NO \\ \hline
		Wadekar et al. \cite{wadekar2019hybrid}                       & AE & CE, KL, ML & OFF & UN &         KEY & ES & OH & SEQ & BIN & THR & NO \\ \hline
		Wan et al. \cite{wan2021glad} (GLAD-PAW)                      & GNN & CE & OFF & UN &                KEY & ES & G & SEQ & PRD & TOP & NO \\ \hline
		Wang et al. \cite{wang2018anomaly}                            & RNN & NA & OFF & SUP &               TOK & TS & SV & OUT & BIN & LAB & NO \\ \hline
		Wang et al. \cite{wang2021log} (CATLog)                       & AE, MLP, TF & CE, CF & OFF & SUP & KEY & ES, TC & SV, DE, CV & SEQ, FREQ & BIN & LAB & NO \\ \hline
		Wang et al. \cite{wang2022lightlog} (LightLog)                & CNN & CE & OFF & SUP &               KEY & ES & SV & SEQ & BIN & LAB & YES \\ \hline
		Wang et al. \cite{wang2021multi} (OC4Seq)                     & RNN & HS & OFF & UN &                KEY & ES & EL & SEQ & TRA & THR & YES \\ \hline
		Wibisono et al. \cite{wibisono2019log}                        & TF, AM & NA & OFF & SEMI &           KEY & ES & OH & SEQ & PRD & TOP & NO \\ \hline
		Wittkopp et al. \cite{wittkopp2021a2log} (A2Log)              & TF, AM & CF & OFF & UN &             TOK & TS & SV & OUT & BIN, TRA & THR & NO \\ \hline
		Xi et al. \cite{xi2021anomaly}                                & RNN, AM & CE & OFF & SEMI &          KEY & ES & SV & SEQ & PRD & TOP & NO \\ \hline
		Xia et al. \cite{xia2021loggan} (LogGAN)                      & RNN, GAN & AT & ON & SEMI &          KEY & ES & OH & SEQ & PRD & THR & NO \\ \hline
		Xiao et al. \cite{xiao2019detecting}                          & RNN, CNN & CE & OFF & SEMI &         KEY & ES & EL & SEQ & PRD & TOP & NO \\ \hline
		Xie et al. \cite{xie2020attention} (ATT-GRU)                  & RNN, AM & MSE & OFF & SEMI &         KEY & ES, PA, EI & PV, EL & SEQ & VEC, PRD & THR, TOP & NO \\ \hline
		Yang et al. \cite{yang2016log}                                & RNN & CE & OFF & SEMI &          TOK & TS & SV & OUT & VEC & THR & NO \\ \hline
		Yang et al. \cite{yang2019nlsalog} (nLSALog)                  & RNN, AM & CE & OFF & SEMI &          KEY & ES & EL & SEQ & PRD & TOP & NO \\ \hline
		Yang et al. \cite{yang2021semi} (PLELog)                      & RNN & NA & OFF & SEMI &              KEY & ES & SV & SEQ & BIN & LAB & NO \\ \hline
		Yen et al. \cite{yen2019causalconvlstm} (CausalConvLSTM)      & RNN, CNN & CE & ON & SEMI &          KEY & ES & OH & SEQ & PRD & TOP & NO \\ \hline
		Yin et al. \cite{yin2020improving} (LogC)                     & RNN & CE & OFF & SEMI &              KEY & ES, PA & OH & SEQ & PRD & TOP & NO \\ \hline
		Yu et al. \cite{yu2021anomaly}                                & RNN & CE & OFF & SEMI &              KEY & ES, EC & SV, ID, CV & SEQ, FREQ & PRD & TOP & NO \\ \hline
		Zhang et al. \cite{zhang2019robust} (LogRobust)               & RNN, AM & CE & OFF & SUP &           KEY & ES & SV & SEQ & BIN & LAB & RE \\ \hline
		Zhang et al. \cite{zhang2021logattn} (LogAttn)                & AE, CNN, AM & CE & OFF & SEMI &      TOK & ES, EC & SV, CV & SEQ, FREQ & RE & THR & NO \\ \hline
		Zhang et al. \cite{zhang2021log} (LSADNET)                    & CNN, TF & CE & ON & SEMI &           KEY & ES, STAT & SV, TM & STAT, SEQ & PRD & TOP & NO \\ \hline
		Zhang et al. \cite{zhang2021sentilog} (SentiLog)              & RNN & NA & OFF & SUP &               TOK & TS & SV & OUT & BIN & LAB & NO \\ \hline
		Zhao et al. \cite{zhao2021trine} (Trine)                      & TF, GAN & NA & OFF & SEMI &          TOK & ES & SV, DE, PE & SEQ & BIN & LAB & NO \\ \hline
		Zhou et al. \cite{zhou2020logsayer} (LogSayer)                & RNN, CNN & NA & ON & SUP &           KEY & EC, STAT & FV & STAT & BIN & LAB & NO \\ \hline
		Zhu et al. \cite{zhu2020approach} (LogNL)                     & RNN & NA & OFF & SEMI &              KEY & ES, PA, EI & SV, PV & SEQ & VEC, PRD & THR & NO \\ \hline
	\end{tabular}
	\label{tab:results}
\end{table*}

\subsection{Deep Learning Techniques}

This section provides an overview of the properties of deep learning models applied in reviewed publications.

\subsubsection{Deep Learning Models} \label{models}

There are many different types of deep learning models (DL-1) that are suitable to be used for anomaly detection in log data \cite{sarker2021deeplearning, sarker2021deepcyber}. The most basic form of a deep learning neural network is that of a \textbf{Multi-Layer Perceptron} (MLP), where all layers in the network are fully connected. Due to their simplicity, their classification accuracies are usually outperformed by other deep learning models that are specifically designed to capture common characteristics present in sequential data. Accordingly, they are rarely considered in the reviewed literature and only occur in combination with other deep learning models or as supplementary attention mechanisms \cite{wang2021log, li2020logspy}. 

\textbf{Convolutional Neural Networks} (CNN) extend upon the architecture of MLPs by inserting convolutional and max pooling layers within the hidden layers. These layers enable that the neural networks capture more abstract features of the input data and at the same time reduce the input dimensions. This has proven especially effective when classifying 2-dimensional input data from images, where features such as lines are learned independent from their exact location in the image. This functionality is transferred to log data by arranging the log keys within a matrix so that the relationships between events, i.e., their temporal dependencies, are captured by the network \cite{lu2018detecting}. There also exist several approaches that rely on specific types of CNNs, such as temporal convolutional networks (TCN) that are specifically designed to process time-series and capture their short- and long-term dependencies through dilated causal convolutions \cite{wang2022lightlog, zhang2021logattn}.

As visible in Table \ref{tab:results}, \textbf{Recurrent Neural Networks} (RNN) are the most commonly used neural network architectures in the surveyed literature, with 36 out of 62 reviewed approaches leveraging RNNs for anomaly detection. The main reason for this is that the architecture of RNNs leverages feedback mechanisms that retain their states over time and thus directly enable learning of sequential event execution patterns in input data, which are the key identifiers for anomalies in log data sets that are commonly used in evaluations (cf. Sect. \ref{eval}). Several different types of RNNs have successfully been applied for this purpose. One of the most widespread architectures are Long Short-Term Memory (LSTM) RNNs that are developed to enable long-time storage of states and comprise cells with input gates, output gates, and forget gates. A majority of the approaches leveraging LSTM RNNs train the network with sequences of event occurrences (or modified and enriched versions thereof) and subsequently disclose unusual sequential patterns in test data as anomalies \cite{du2017deeplog}. Some approaches make use of Bi-LSTM RNNs, which are basically two independent LSTM RNNs that work in parallel and process sequences in opposite directions, i.e., while one LSTM RNN processes the input sequences as usual from the first to the last element, the other LSTM RNN processes sequence elements starting from the last entry and predicts elements that chronologically precede them. Experiments suggest that Bi-LSTM RNNs outperform LSTM RNNs \cite{zhang2019robust, ott2021robust, li2020swisslog, farzad2019log, zhang2021sentilog, syngal2021server, yu2021anomaly, sun2020context}. Another popular choice for RNNs are Gated Recurrent Units (GRU) that simplify the cell architecture as they only rely on update and reset gates. One of the main benefits of GRUs is that they are computationally more efficient than LSTM RNNs, which is a relevant aspect for use cases focusing on edge devices \cite{xie2020attention, wang2021multi, farzad2020log, farzad2019log, sundqvist2020boosted, yang2021semi, studiawan2021anomaly, gu2021research, yang2016log}.

While aforementioned deep learning models are primarily used for classification problems across different research fields, there are also models that are specifically designed to operate in unsupervised manner and are thus a natural choice for anomaly detection. One of them are \textbf{Autoencoders} (AE), which first create a code from the input data using an encoder and then try to approximate the input from the code using a decoder, thereby avoiding the need for labeled input data. The main idea is that through this process the neural network learns the main features from the input but neglects the noise in the data, similar to dimension reduction techniques such as principal component analysis (PCA). Any input data that is fed into an already trained network and yields a high reconstruction error is then considered as anomalous. Besides the standard model for Autoencoders, there are also several related types, such as Variational Autoencoders (VAE) that operate on statistical distributions \cite{wadekar2019hybrid, qian2020anomaly, wang2021log}, Conditional Variational Autoencoders (CVAE) that add conditional information such as event types to the training \cite{otomo2019latent}, and Convolutional Autoencoder (CAE) that leverage the advantages of CNNs regarding learning of location-independent features \cite{wadekar2019hybrid}.

\textbf{Generative Adversarial Networks} (GAN) are another approach for unsupervised deep learning. They actually consist of two separate components that compete with each other: a generator that produces new data that resembles the input data, and a discriminator that estimates the probability that some data stems from the input data, which is used to improve the generator. Existing approaches use different models to construct GANs, including LSTM RNNs \cite{xia2021loggan, han2021unsupervised}, CNNs in combination with GRUs \cite{farzad2020log}, and Transformers \cite{zhao2021trine, wang2021log}.

\textbf{Transformers} (TF) make use of so-called self-attention mechanisms to embed data instances into a vector space, where similar instances should be closer to each other than dissimilar ones \cite{nedelkoski2020self, guo2021logbert, du2021log}. The goal of Transformers is to assign weights to specific inputs according to the context of their occurrence, such as words in sentences. Accordingly, Transformers have been particularly successful in the area of natural language processing (NLP). \textbf{Attention mechanisms} (AT) do not just appear in Transformers, but are also frequently used to improve classification and detection in other deep neural networks such as RNNs by weighting relevant inputs higher. This effect is particularly strong when long sequences are ingested by RNNs \cite{zhang2021logattn}. Attention mechanisms are usually realized as trainable networks such as MLPs \cite{li2020logspy}. In order to avoid confusions with the Transformer model, we state these additional attention mechanisms explicitly in Table \ref{tab:results}.

Wan et al. \cite{wan2021glad} are the only authors to utilize \textbf{Graph Neural Networks} (GNN). While neural networks typically ingest ordered data, e.g., CNNs rely on 2-dimensional input data and RNNs require sequences of observations, GNNs are designed to ingest graph inputs, i.e., sets of vertices and edges. One possibility to transform log data into graphs is to generate session graphs, with vertices representing events and edges their sequential executions. Another less commonly applied type of deep learning model is the \textbf{Evolving Granular Neural Network} (EGNN), a fuzzy inference system that is gradually constructed from online data streams \cite{decker2020comparison}.

Our survey shows that many of the reviewed approaches rely on only one type of deep learning model. However, some authors also use combinations of different models. For example, Wang et al. \cite{wang2021log} propose to use a MLP to combine the output of a VAE with that of a Transformer trained with adversarial learning.

\subsubsection{Training loss function}

\begin{table*}[]
	\footnotesize
	\centering
	\caption{Definitions of common training loss functions}
	\begin{tabular}{llc}
		\textbf{Name} & \textbf{Description} & \textbf{Equation} \\ \hline \hline
		Cross-Entropy & Loss between a ground truth label $y$ and the predicted label $\hat{y}$. & $H(y, \hat{y}) = - \sum_{j=1}^{N} y_j log\left( \hat{y}_j \right)  $ \\ \hline 
		Hyper-Sphere Objective Function & Distance between embedding vector $y$ and hyper-sphere center $c$. & $L_{HS} = \frac{1}{N} \sum_{j=1}^{N} \left\| y_j - c \right\|^2 $ \\ \hline 
		Mean Squared Error & Loss between a ground truth label $y$ and the predicted label $\hat{y}$. & $L_{MSE}(y, \hat{y}) = \frac{1}{N} \sum_{j=1}^{N} \left( y_j - \hat{y}_j \right)^2 $ \\ \hline 
		Kullback-Leibler Divergence & Statistical divergence between probability distributions $P$ and $Q$. & $KL(P \| Q) = \sum_{x \in \mathcal{X}} P(x) log\left( \frac{P(x)}{Q(x)} \right) $ \\ \hline 
		Adversarial Training Function & Loss for vector $y$ and prediction $\hat{y}$ with generator $G$ and discriminator $D$. & $\!\begin{aligned} L_{AT} = & \min_G \max_D ( \mathbb{E}\left( log\left( D(G(\hat{y})) \right)  \right) + \\
			& \mathbb{E}\left( log\left( 1 - D(G(y)) \right)  \right) ) \end{aligned} $ \\ \hline
	\end{tabular}
	\label{tab:loss}
\end{table*}

Loss functions (DL-2) are essential for training of deep neural networks as they quantify the difference between the output of the neural network and the expected result \cite{zhang2018generalized}. The most common loss function in the reviewed publications is the \textbf{Cross-Entropy} (CE), in particular, the categorical cross-entropy for multi-class prediction \cite{du2017deeplog, syngal2021server} or binary cross-entropy that only differentiates between the normal and anomalous class \cite{wang2022lightlog}. Other common loss functions include the \textbf{Hyper-Sphere Objective Function} (HS) where the distance to the center of a hyper-sphere represents the anomaly score \cite{nedelkoski2020self, wang2021multi, guo2021logbert, han2021unsupervised}, the \textbf{Mean Squared Error} (MSE) that is used for regression \cite{bursic2019anomaly, ott2021robust, catillo2022autolog, liu2021lognads, studiawan2021anomaly, xie2020attention, du2017deeplog}, and the \textbf{Kullback-Leibler Divergence} (KL) and its extension \textbf{Marginal Likelihood} (ML) that are useful to measure loss in probability distributions \cite{otomo2019latent, wadekar2019hybrid}.

Some of the presented approaches are trained with \textbf{Custom Loss Functions} (CF), including combinations of CE and HS \cite{wittkopp2021a2log}. Some authors also define objective functions specifically for \textbf{Adversarial Training} (AT) of GANs \cite{xia2021loggan, han2021unsupervised}. Table \ref{tab:loss} summarizes the main loss functions. Out of all publications, 14 do not state the loss function and are therefore marked as \textbf{Not Available} (NA).

\subsubsection{Operation mode}

When applying anomaly detection in real world scenarios, not all log data is available at any time; instead, events are generated as a continuous stream and should be analyzed only at their time of occurrence in order to enable (close to) real-time detection. Thereby, the structural integrity and statistical properties of the generated logs vary over time, e.g. the overall system utilization is not stationary as user interactions and the technological environment are subject to change. In addition, log templates change or new log events occur when applications generating these logs are modified \cite{zhang2019robust}.

To keep up with these changes in an automated way, algorithms need to adopt online or incremental learning (DL-3), i.e., ingest input data with linear time complexity so that data sets are processed in a single pass where each data instance is only handled once. While it is usually always possible to carry out detection in an online fashion when trained models are considered static, it is significantly more challenging to develop algorithms that support online learning and dynamically update their models to adapt to new events or patterns by incorporating them into the baseline for detection \cite{cui2016continuous, hadsell2020embracing}. As continuous learning is still an open problem, we mark all approaches that enable dynamic model updates at least to some degree as \textbf{online} (ON) and all other approaches with offline training phases as \textbf{offline} (OFF) in Table \ref{tab:results}.

The reviewed approaches addressed dynamic model adaptation in multiple ways. Du et al. \cite{du2017deeplog} suggest to update the weights of their neural network when false positives are identified during the detection to reflect the correct event probability distributions without the need to re-train from scratch. Meng et al. \cite{meng2019loganomaly} argue that such a manual feedback loop is infeasible in practice and therefore resort to re-training where new event types are mapped to existing ones. Yen et al. \cite{yen2019causalconvlstm} automatically re-train their neural network on batches of new log data when false positive rates exceed a certain threshold. However, calculating the false positive rates relies on labeled data and thus does not remove the human from the loop. A promising solution to aforementioned problems is presented by Decker et al. \cite{decker2020comparison}, who employ an evolving classifier that is specifically designed to handle unstable data streams and could thus enable continuous learning. 

In general, both online and offline models can work in \textbf{supervised} (SUP) as well as \textbf{unsupervised} (UN) fashion (DL-4). However, we noticed that almost all supervised learning approaches in the reviewed publications opt for an offline training phase. This is reasonable as the label information required for supervised learning relies on manual analysis or validation and therefore can only be generated forensically for delimited data sets, but not for data streams.

We also made the observation that many reviewed approaches claim to enable unsupervised learning, but are in fact operated in \textbf{semi-supervised} (SEMI) fashion as they usually assume that training takes place only on normal data that is free of anomalies \cite{chenunsupervised}. The main problem is that anomalies that are present in training data would incorrectly change the weights of the neural networks and thus deteriorate their detection in the subsequent detection phase. Accordingly, only deep learning models that are designed for unsupervised learning are capable of handling anomalies in training data, for example, the approach based on a CVAE model presented by Otomo et al. \cite{otomo2019latent}. Note that neural network architectures that would support unsupervised learning may also be applied for semi-supervised detection and are therefore not necessarily marked as unsupervised.

\subsection{Log Data Preparation} \label{preproc}

\begin{figure*}
	\centering
	\includegraphics[width=0.99\textwidth]{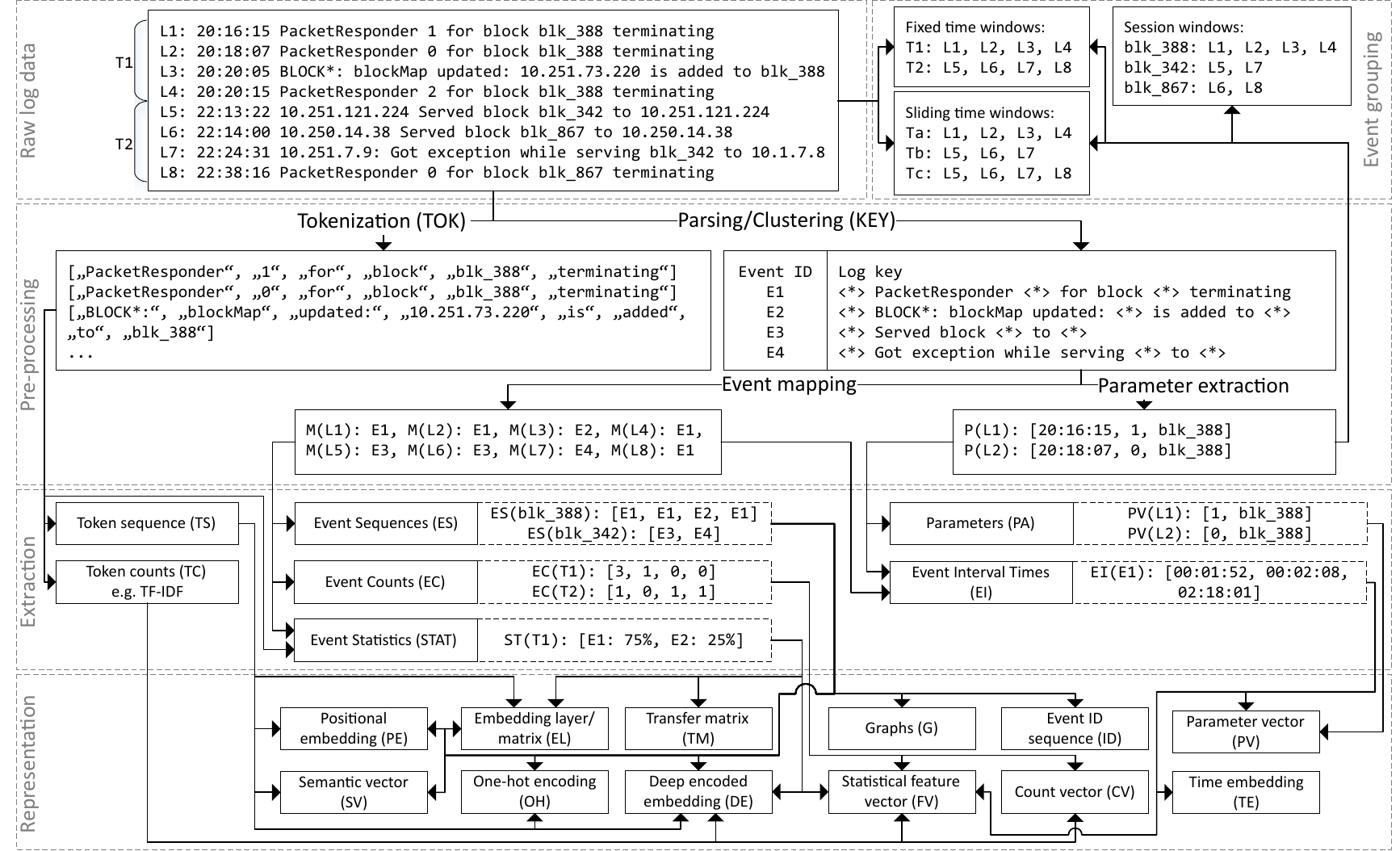}
	\caption{Different stages of data preparation to transform raw log data into numeric vectors.}
	\label{fig:preprocessing}
\end{figure*}

This section outlines all steps necessary to prepare raw and textual log data for deep learning. This includes grouping of events into windows or sessions, tokenization and log parsing, extraction of features from the logs, as well as transformation of these features into vector representations that are suitable as input to neural networks. Figure \ref{fig:preprocessing} provides an overview of these data preparation methods used in the reviewed literature and illustrates them on the sample log lines L1-L8. In the following sections, we discuss all stated methods in detail.

\subsubsection{Pre-processing} \label{preprocessing}

As log data is generally unstructured it is necessary to pre-process them in some way before feeding them into deep learning systems (PP-1). Our survey shows that there are two main approaches to handle unstructured data. The most common approach is to leverage parsers, which are usually referred to as \textbf{log keys} (KEY), to extract unique event identifiers for each line as well as event parameter values as described in Sect. \ref{logdatadef}. Thereby, authors usually re-use existing log keys for well-known data sets or create the keys using state-of-the-art approaches for parser generation, such as Drain \cite{he2017drain} or Spell \cite{du2016spell}. Typically, each line matches exactly one key; it is thus easy to assign a unique event type identifier to each log line during event mapping. For example, in Fig. \ref{fig:preprocessing} line L1 is mapped to event type identifier E1 as it matches the corresponding log key. Moreover, the figure also shows that parsing allows to extract all parameters from log events, in particular, the time stamp and identifiers for packet responder and file block are extracted from L1 and stored in a list.

An alternative to parsing are \textbf{token-based} (TOK) strategies that split log messages into lists of words, for example, by splitting them at white spaces. It is then common to clean the data by transforming all letters to lowercase and removing special characters and stop words before obtaining the final word vectors. While such approaches draw less semantic information from the single tokens, they have the advantage of being more flexible as they rely on generally applicable heuristics rather than pre-defined parsers and are therefore widely applicable. Some approaches make use of a \textbf{combination} (COM) of parsing and token-based pre-processing strategies, in particular, by generating token vectors from parsed events rather than raw log lines \cite{le2021log, catillo2022autolog, guo2021anomaly}.

\subsubsection{Event grouping} \label{eventgrouping}

As pointed out in Sect. \ref{anomalydetection}, simple outlier detection does not require any grouping of logs since single unusual events are regarded as anomalies independent from the context in which they occur in. However, deep learning is most often applied to disclose unusual patterns of multiple log events, such as changes of event sequences or temporal log correlations. For these cases it is necessary to logically organize events into groups that are then analyzed individually or in relation to each other.

We illustrate common event grouping strategies in the top right of Fig. \ref{fig:preprocessing}. Grouping into time windows is almost always feasible as log events are generally produced with a time stamp that documents their time of generation \cite{landauer2020system}. Since time stamps commonly occur in the beginning of log lines and are thus relatively easy to extract, it is not necessary to parse the remainder of the usually more complex log messages. There are two main strategies for time-based grouping \cite{he2016experience}. First, \textbf{sliding time windows} are windows of a specific duration that are shifted across the log data with a fixed step size that is generally smaller and a whole number divisor of the window size. For every step where the time window is moved all logs with a time stamp within the current start and end time of the window are allocated to the same group, where each log line may appear in multiple groups as time windows overlap. For the sample logs in Fig. \ref{fig:preprocessing} we assume time window of 2 hours starting at 20:00:00 and a step size of 30 minutes. As visible in the figure, lines L5, L6, and L7 are contained in both $Tb$ and $Tc$. Second, \textbf{fixed time windows} are special cases of sliding time windows where the step size is set to the same value as the window size. While this strategy results in a less fine-grained view on the data, the advantage is that each log event will only be allocated to exactly one time window, which makes subsequent computations such as time-series analysis easier. The log events in Fig. \ref{fig:preprocessing} are exemplarily grouped into fixed time windows of 2 hours, yielding $T1$ containing the first set of four lines and $T2$ containing the subsequent set of four lines. It must be noted that grouping based on sliding or fixed windows may also be carried out by numbers of lines rather than time so that each group of lines has the same size. While this avoids the need to process time stamps and ensures that the group sizes are fixed (e.g., there cannot be any empty groups), it is more difficult to consider event frequencies as time-series as the resulting windows represent varying time spans.

An entirely different grouping strategy are \textbf{session windows} that rely on an event parameter that acts as an identifier for a specific task or process where the event originated from. Grouping log events by these session identifiers allows to extract event sequences that correctly depict underlying program workflows even when multiple sessions are carried out in parallel on the monitored system. Unfortunately, not all types of log data come with such an identifier for sessions. In the sample logs depicted in Fig. \ref{fig:preprocessing} the file block identifier (e.g., blk\_388) acts as a session identifier and thus allows us to exemplarily extract three event sequences.

\subsubsection{Feature extraction}

The parsing- and token-based pre-processing strategies described in Sect. \ref{preprocessing} enable the extraction of structured features from the otherwise unstructured logs (PP-2). Some of these features are directly derived from pre-processing logs, e.g., the \textbf{Token Sequence} (TS) may be used without any further modifications for analyzing each log line as a sentence of words. On the other hand, \textbf{Token Counts} (TC) require an additional step of computation where the tokens in each line are compared and counted, including advanced weighting techniques for each token such as term frequency–inverse document frequency (TF-IDF) that estimates token relevance based on token occurrences across all observed log lines.

Considering the outcome of the event mapping step it is simple to extract \textbf{Event Sequences} (ES), i.e., sequences of event type identifiers that are usually separated by fixed, sliding, or session windows (cf. Sect. \ref{eventgrouping}). For example, Fig. \ref{fig:preprocessing} shows that the event sequence for file block blk\_388 is $\left[ E1, E1, E2, E1 \right]  $ corresponding to the event identifiers for the respective log keys matching the lines L1-L4. \textbf{Event Counts} (EC) are vectors of length $d$, where the $i$-th element of the vector depicts the number of occurrences of the $i$-th log key and $d$ is the total number of available log keys. The example in Fig. \ref{fig:preprocessing} shows the event count vector for time window $T1$ as $\left[ 3, 1, 0, 0 \right]  $, indicating that three log lines corresponding to the first log key (E1) appeared in $T1$, in particular, the lines L1, L2, and L4. Besides frequencies, many other \textbf{statistical properties} (STAT) may be computed from event occurrences, such as the percentage of seasonal logs \cite{zhou2020logsayer}, the lengths of log messages \cite{studiawan2021anomaly}, log activity rates \cite{decker2020comparison}, entropy-based scores for chunks of log lines \cite{catillo2022autolog}, or the presence of sudden bursts in event occurrences \cite{zhou2020logsayer}.

\textbf{Parameters} (PA) of log events are extracted as lists of values. Since the semantic meaning of each parameter is known after parsing, the values in each vector can be analyzed with methods that are appropriate for the respective value types, e.g., numeric or categorical. One special parameter of log events is the time stamp as it allows to put event occurrences into chronological order and infer dynamic dependencies. Accordingly, \textbf{Event Interval Times} (EI), i.e., inter-arrival times of log lines that belong to the same event type, are a frequently extracted feature for anomaly detection. 

\subsubsection{Feature representation} \label{featurerepr}

The extracted features described in the previous section comprise numeric or categorical vectors and are suitable to be consumed by neural networks. For example, event sequences are represented as \textbf{Event ID sequence vectors} (ID), i.e., chronologically ordered sequences of log key identifiers, and fed into RNNs to learn dependencies of event occurrences and disclose unusual sequence patterns as anomalies \cite{sundqvist2020boosted}. Event counts are represented as ordered \textbf{Count Vectors} (CV) and are also similarly used as input for RNNs \cite{baril2020application}. Event statistics are another type of input that do not require any specific processing other than representing them in a \textbf{Statistical Feature Vector} (FV) where the position of each element in the vector corresponds to one particular feature.

While it is possible to directly use the extracted features, most of the approaches presented in the reviewed publications in fact rely on combinations or otherwise transformed vector representations of the original feature vectors (PP-3). Thereby, the most common representation is that of a \textbf{Semantic Vector} (SV). Within the field of NLP it is common practice to transform words of a sentence into so called semantic vectors that encode context-based semantics (e.g. Word2Vec \cite{mikolov2013efficient}, BERT \cite{devlin2018bert}, or GloVe \cite{pennington2014glove}) or language statistics (e.g. TF-IDF \cite{manning2010introduction}). Since each log line comprises sequences of tokens analogous to words in a sentence, it stands to reason to apply methods from natural language processing on the token sequences. Similarly, a sequence of multiple subsequent events can be regarded as a sentence, where each unique log key represents a word. Semantic encoding is typically achieved by training deep neural methods on a specific log file or by relying on pre-trained models. Semantic vectors are sometimes used in combination with \textbf{Positional Embedding} (PE), where elements (typically tokens) are encoded based on their relative positions in a sequence. To add the positional information to the encoded log messages authors usually use sine and cosine functions for even and odd token indices respectively \cite{le2021log, nedelkoski2020self, zhao2021trine}. 

\textbf{One-Hot Encoding} (OH) is one of the most common techniques to handle categorical data and is therefore frequently applied on event types (as a finite number of log keys is defined in the parser) or token values. Formally, the one-hot encoding of a value $i$ from an ordered list of $d$ values is a vector of length $d$ where the $i$-th element is $1$ and all others are $0$ \cite{du2017deeplog}. While most authors use one-hot encoded data directly as an input to neural networks, it is also possible to combine it with other features such as count vectors so that the applied neural network is capable of discerning the input and learn separate models for different log keys \cite{otomo2019latent}.

\textbf{Embedding Layers/Matrices} (EL) are typically used to resolve the problems with respect to sparsity of high-dimensional input data such as one-hot encoded event types when a large number of log keys are required to parse the logs \cite{wang2021multi, yang2019nlsalog}. They are usually randomly initialized parameters which are trained alongside the classification models to create optimal vector encodings for log messages \cite{lu2018detecting}. The vector encodings are usually arranged in a matrix so that the respective vector for a particular log key is obtained by multiplying the matrix with a one-hot encoded log key vector. The main difference to semantic vectors is that embedding layers/matrices are generally not trained towards NLP objectives, i.e., they do not aim to learn the semantics of words like Word2Vec; instead, embedding layers/matrices are only trained to minimize the loss function of the classification network. Some authors also use custom embedding models based on deep learning; we refer to their output as \textbf{Deep Encoded Embeddings} (DE). This includes a combination of character-, event- and sequence-based embeddings \cite{hashemi2021onelog}, attention mechanisms using MLPs and CNNs \cite{li2020logspy}, and token counts with label information fed into VAEs \cite{wang2021log}. 

Other than aforementioned methods that operate with event types and mostly use tokens only for encoding these events in vector format, approaches that rely on \textbf{Parameter Vectors} (PV) directly use the actual values extracted from the parsed log messages. Thereby, extracted parameters that are numeric may be used for multi-variate time-series analysis with RNNs \cite{xie2020attention, zhu2020approach, du2017deeplog}, while categorical values could be one-hot encoded and vectorized with word embedding methods \cite{huang2020hitanomaly}. Either way, as different log events have varying numbers of parameters with different semantic meaning, it is usually necessary to analyze the parameters of each event type on their own \cite{xie2020attention, zhu2020approach, du2017deeplog}. The time stamp of log events is a special parameter as it allows to put other parameters in temporal context, which is required for time-series analysis. However, the time stamps themselves may be used for \textbf{Time Embedding} (TE) and serve as input to neural networks \cite{bursic2019anomaly}. For this purpose, Li et al. \cite{li2020swisslog} generate vectors for sequences of time differences between event occurrences by applying soft one-hot encoding.

While aforementioned representations are used in different variations in several publications, \textbf{Graphs} (G) are an entirely different approach that is only employed by Wan et al. \cite{wan2021glad}. The key idea is to transform event sequences into session graphs and then apply neural networks that are specifically designed for these data structures. Another less common strategy for encoding dependencies between log messages is the so-called \textbf{Transfer Matrix} (TM). In particular, the $d \times d$-dimensional matrix encodes the probabilities that any of the $d$ log keys follows another \cite{zhang2021log}.

\subsection{Anomaly Detection Techniques}

The previous sections described methods to prepare the input log data for ingestion by neural networks. However, in many cases it is non-trivial to retrieve the information whether a data sample presented to the neural network is anomalous or not as there are no dedicated output nodes for anomalies that are not known beforehand. This section therefore outlines the anomaly detection techniques applied by the reviewed approaches. First, we state different types of anomalies that are commonly targeted by approaches. Second, we investigate how the detection or classification result is retrieved from the network output. Finally, we state different decision criteria that are used to differentiate normal from anomalous samples based on these resulting measures.

\subsubsection{Anomaly types}

The reviewed approaches address different types of anomalies as outlined in Sect. \ref{anomalydetection} (AD-1). \textbf{Outliers} (OUT) are single log events that do not fit to the overall structure of the data set. Most commonly, outlier events are detected based on their unusual parameter values \cite{hirakawa2020software}, token sequences \cite{wang2018anomaly, wittkopp2021a2log}, or occurrence times \cite{bursic2019anomaly}. Comparatively few approaches consider outliers since the majority of reviewed approaches focus on collective anomalies, in particular, involving sequences of events. 

\textbf{Sequential} (SEQ) anomalies are detected when execution paths change, i.e., the applications that generate logs execute events differently than before. This could result in additional, missing, or differently ordered events within otherwise normal event sequences as well as completely new sequences that could even involve previously unseen event types. A common method to detect these anomalies is to check whether a specific event type in a sequence of events is expected to occur given all the events that occur before or thereafter.

While the detection of sequential anomalies inherently assumes that events occur as ordered sequences, \textbf{frequency} (FREQ) anomalies only consider the number of event occurrences. Nonetheless, event frequencies may be used to infer dependencies between events, for example, the numbers of events related to opening and closing files should be the same as every file will eventually be closed and needs to be opened before doing so \cite{meng2019loganomaly}. The main idea for detecting frequency anomalies is that changes of system behavior affect the number of usual event occurrences that are most often counted within time windows. 

Some approaches also consider anomalies that base on certain quantitatively expressed properties of multiple log events that go beyond event counts, such as their inter-arrival times \cite{li2020swisslog} or seasonal occurrence patterns \cite{zhou2020logsayer}. We refer to them as \textbf{statistical} (STAT) anomalies, because their detection generally assumes that the event occurrences follow specific stable distributions over time. The following section describes how the output of the neural networks is used to determine the aforementioned types of anomalies.

\subsubsection{Network output}

In general, the output of a neural network consists either of a single node or multiple nodes in its final layer (AD-2). Accordingly, the resulting value extracted from the network is a scalar or vector of numeric values. One possibility is to consider these results as an anomaly score that expresses to what degree the log events presented to the network represent an anomaly or not. As these scores are generally difficult to interpret on their own, it is usually necessary to compare them with some threshold. In \textbf{binary classification} (BIN) this idea is used to estimate whether the input presented to the neural network is either normal or anomalous. For supervised approaches the numeric output can be interpreted as probabilities that the input corresponds to either class. On the other hand, anomaly scores that are generated by semi- or unsupervised approaches are generally not normalized, e.g., the distance between the input and the center of a hyper-spherical cluster can become arbitrarily large, and therefore need to be compared to empirically determined thresholds. Similarly, \textbf{Input vector transformations} (TRA) that transform the input into a new vector space and generate clusters for normal data are capable of detecting outliers by their large distances to cluster centers. Another related method is to leverage the \textbf{reconstruction error} (RE) of Autoencoders that first encode the input data in a lower dimensional space and then attempt to reconstruct them to their original form. In this case, input samples are considered anomalous if they are difficult to reconstruct, i.e., yield a large reconstruction error, because they do not correspond to the normal data that the network was trained on.

While aforementioned approaches pursue binary classification that separates normal from anomalous input, there are also concepts that are capable of differentiating between more than two classes. \textbf{Multi-class classification} (MC) assigns distinct labels to specific types of anomalies but requires supervised learning in order to capture the patterns specific to these classes in the training phase. To resolve this issue, it is also possible to consider event types as the target of classification. The most common approach for this is to train the models to predict the next log key following a sequence of observed log events. When a softmax function is used as an activation for the output, this prediction yields a \textbf{probability distribution} (PRD) with the individual probabilities for each log key. The problem of predicting the next log event in a sequence can also be formulated as a regression task when events are considered as \textbf{numeric vectors} (VEC), in particular, semantic or parameter vectors. Thereby, the neural network outputs a vector representing the expected event vector instead of the respective event type.

\subsubsection{Detection method}

The different strategies for obtaining the network output described in the previous section already give a rough idea on the methods used to differentiate normal from anomalous behavior and eventually report anomalies (AD-3). When the network output directly corresponds to a particular \textbf{label} (LAB), for example, as accomplished by binary classification, it is simple to generate anomalies for all samples that are labeled as anomalous. For all approaches that output some kind of numeric value or anomaly score it is straightforward to use a \textbf{threshold} (THR) for differentiation. This threshold is also useful to tune the detection performance of the approach and find an acceptable tradeoff between TPR and FPR by empirical experimentation. Another approach is to model the anomaly scores obtained from the network as statistical distributions. In particular, Du et al. \cite{du2017deeplog} and Xie et al. \cite{xie2020attention} use a Gaussian distribution to detect parameter vectors with errors outside of specific confidence intervals as anomalous. A different approach is proposed by Otomo et al. \cite{otomo2019latent}, who apply clustering on the reconstruction errors and detect all samples that are sufficiently far away from the normal clusters as anomalies.

\begin{figure}
	\centering
	\includegraphics[width=1\columnwidth]{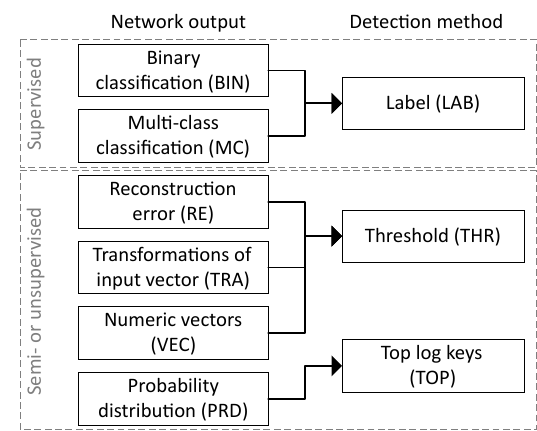}
	\caption{Most common combinations of network output types and detection techniques.}
	\label{fig:ad}
\end{figure}

When the output of the neural network is a multi-class probability distribution for known log keys it is common to consider the top $n$ log keys with the \textbf{highest probabilities} (TOP) as candidates for classification. Thereby, an anomaly is only detected if the actual type of the log event is not within the set of candidates. The number of considered candidates $n$ regulates the tradeoff between TPR and FPR analogous to the aforementioned threshold. Figure \ref{fig:ad} shows how the network output relates to the applied detection techniques. Both BIN and MC rely on supervised learning and are therefore able to directly assign labels to new input samples. All other techniques enable semi- or unsupervised training. In particular, RE, TRA, and VEC produce anomaly scores that are compared against thresholds, while PRD is typically compared against the top log keys with highest probabilities. Note that there are some exceptions to this overall pattern. For example, the approach by Yang et al. \cite{yang2021semi} supports semi-supervised training through the use of probabilistic labels, and the approaches by Zhang et al. \cite{zhang2019robust} and Syngal et al. \cite{syngal2021server} are supervised despite relying on reconstruction errors.

\subsection{Evaluation \& Reproducibility} \label{eval}

This section provides an overview of evaluations carried out in the reviewed publications. We present a list of openly available data sets that are useful for evaluations and state commonly used evaluation metrics and benchmarks. We also discuss reproducibility of evaluations with respect to the availability of open-source implementations.

\subsubsection{Data sets} \label{datasets}

Data sets are essential in scientific publications to validate the approach and show improvements over state-of-the-art detection rates (ER-1). Our literature review reveals that there are only few data sets that are commonly used when evaluating log data anomaly detection approaches using deep learning. Table \ref{tab:eval} shows an overview of all data sets used in the reviewed publications. As visible in the table, the vast majority of evaluations rely on only four data sets (HDFS, BGL, Thunderbird, and OpenStack). In the following, we briefly describe each data set.

\begin{table*}[]
	\small
	\caption{Common public log data sets}
	\begin{tabular}{lcccccp{0.25\textwidth}}
		\textbf{Data set} & \textbf{Year} & \textbf{Use-case} & \textbf{Labels} & \textbf{Sessions} & \textbf{Anomaly sources} & \textbf{Used in evaluation} \\ \hline \hline
		HDFS \cite{xu2009detecting} & 2009 & High-perf. comp. & \checkmark & \checkmark & \makecell[ct]{Failures} & \cite{zhang2019robust, zhang2021logattn, wang2022lightlog, bursic2019anomaly, lu2018detecting, zhou2020logsayer, du2017deeplog, chenunsupervised, meng2019loganomaly, lv2021conanomaly, guo2021logbert, yang2019nlsalog, wang2021log, li2020swisslog, xie2020attention, wang2021multi, sundqvist2020boosted, zhu2020approach, yang2021semi, xi2021anomaly, patil2019explainable, cheansunan2019detecting, qian2020anomaly, gu2021research, guo2021anomaly, yu2021anomaly, yin2020improving, liu2021lognads, guo2021translog, hashemi2021onelog, yen2019causalconvlstm, wadekar2019hybrid, du2021log, wan2021glad, chen2020logtransfer, zhang2021log, xia2021loggan, sun2020context, huang2020hitanomaly, zhao2021trine, xiao2019detecting, le2021log} \\ \hline 
		BlueGene/L (BGL) \cite{oliner2007supercomputers} & 2007 & High-perf. comp. & \checkmark & - & Failures & \cite{zhang2021logattn, wang2022lightlog, du2017deeplog, chenunsupervised, meng2019loganomaly, wittkopp2021a2log, lv2021conanomaly, guo2021logbert, farzad2021two, yang2019nlsalog, wang2021log, han2021unsupervised, li2020swisslog, wang2021multi, nedelkoski2020self, farzad2019log, farzad2020log, yang2021semi, xi2021anomaly, studiawan2020anomaly, gu2021research, catillo2022autolog, liu2021lognads, guo2021translog, hashemi2021onelog, du2021log, wan2021glad, zhang2021log, xia2021loggan, sun2020context, huang2020hitanomaly, hirakawa2020software, le2021log} \\ \hline
		Thunderbird \cite{oliner2007supercomputers} & 2007 & High-perf. comp. & \checkmark & - & Failures & \cite{zhang2021logattn, wittkopp2021a2log, guo2021logbert, farzad2021two, han2021unsupervised, nedelkoski2020self, farzad2019log, farzad2020log, wang2018anomaly, guo2021anomaly, yin2020improving, guo2021translog, sun2020context, le2021log} \\ \hline
		OpenStack \cite{du2017deeplog} & 2017 & Virtual machines & \checkmark & \checkmark & \makecell[ct]{Failures} & \cite{zhou2020logsayer, du2017deeplog, baril2020application, farzad2021two, ott2021robust, farzad2019log, farzad2020log, zhu2020approach, qian2020anomaly, hashemi2021onelog, wibisono2019log, huang2020hitanomaly, zhao2021trine} \\ \hline 
		Hadoop \cite{lin2016log} & 2016 & High-perf. comp. & \checkmark & \checkmark & \makecell[ct]{Failures} &  \cite{catillo2022autolog, hashemi2021onelog, chen2020logtransfer, studiawan2020anomaly} \\ \hline
		Spirit \cite{oliner2007supercomputers} & 2007 & High-perf. comp. & \checkmark & - & Failures & \cite{wittkopp2021a2log, nedelkoski2020self, le2021log} \\ \hline
		Digital Corpora \cite{garfinkel2009bringing} & 2009 & OS logs & - & - & Failures & \cite{studiawan2020anomaly, studiawan2021anomaly} \\ \hline
		DFRWS 2009 \cite{dfrws2009} & 2009 & OS logs & - & - & Exfiltration & \cite{studiawan2020anomaly, studiawan2021anomaly} \\ \hline
		Honeynet 2011 \cite{honeynet2011} & 2011 & OS logs & - & - & Intrusions & \cite{studiawan2020anomaly, studiawan2021anomaly} \\ \hline
		Windows \cite{he2020loghub} & 2020 & OS logs & - & - & - & \cite{studiawan2020anomaly, studiawan2021anomaly} \\ \hline
		Linux Security Logs \cite{chuvakin} & 2020 & OS logs & - & - & Intrusions & \cite{studiawan2021anomaly} \\ \hline
		Honeynet 2010 \cite{honeynet2010} & 2010 & OS logs & - & - & Intrusions & \cite{studiawan2020anomaly} \\ \hline
		Android \cite{he2020loghub} & 2020 & Mobile OS logs & - & - & - & \cite{li2020swisslog} \\ \hline
		HPC RAS \cite{zheng2011co} & 2011 & High-perf. comp. & - & \checkmark & Failures & \cite{nedelkoski2020self} \\ \hline
		Spark \cite{he2020loghub} & 2020 & High-perf. comp. & - & - & Failures & \cite{studiawan2020anomaly} \\ \hline
		Zookeeper \cite{he2020loghub} & 2007 & High-perf. comp. & - & - & Failures & \cite{studiawan2020anomaly} \\ \hline
	\end{tabular}
	\label{tab:eval}
\end{table*}

The \textbf{HDFS} log data set stems from the Hadoop Distributed File System (HDFS) running on a high-performance computing cluster with 203 nodes that computes many standard MapReduce jobs. More than 24 million logs are collected over a period of two days. The data set comprises sequences of heterogeneous log events for specific file blocks that act as identifiers for sessions. Some of the event sequences correspond to anomalous execution paths that are mostly related to performance issues such as write exceptions, which were manually labeled \cite{xu2009largescale}. The sample logs shown in Fig. \ref{fig:preprocessing} are simplified versions of the logs from the HDFS data set.

The \textbf{BGL} data set comprises more than four million log events that were collected over more than 200 days from a BlueGene/L (BGL) supercomputer running at the Lawrence Livermore National Labs. The log events were generated with a severity field that allows to separate them into classes; however, the logs were additionally labeled manually by system administrators. The anomalies occurring in the logs correspond to both hardware and software problems. Other log data sets from the same family that are also presented in the study by Oliner et al. \cite{oliner2007supercomputers} are \textbf{Thunderbird} and \textbf{Spirit}. These data sets comprise system logs (syslog) and similarly comprise alerts related to system problems such as disk failures. The events in both data sets include automatically generated alert tags that can be used as labels. The \textbf{HPC RAS} data set \cite{zheng2011co} is another log data set from the same family of supercomputers. Reliability, Availability, and Serviceability (RAS) logs are usually used to understand the reasons for system failure. However, this log data set does not include labels for anomalies. 

The \textbf{OpenStack} data set was collected from an OpenStack platform where automatic scripts continuously and randomly carry out tasks related to handling of virtual machines, including creation, pausing, deletion, etc. Other than aforementioned data sets that mostly comprise randomly occurring failures, the authors of the OpenStack data set purposefully injected anomalies at specific points in time, including timeouts and errors. The events include instance identifiers that can be used to identify sessions \cite{du2017deeplog}.

Similar to the HDFS data, the \textbf{Hadoop} data set comprises logs from a computing cluster that runs the MapReduce jobs WordCount and PageRank. After an initial training phase, the authors purposefully trigger failures on the nodes, in particular, by shutting down a machine, disconnecting the network, and filling up the hard disk of one server. The logs are separated into different files according to application identifiers that act similar to session identifiers and are also used by the authors to assign labels to anomalous program executions \cite{lin2016log}.

Loghub \cite{he2020loghub} comprises several data sets that are used in evaluations, including logs from high-performance computing systems. The \textbf{Spark} data set contains logs from the distributed data processing engine Apache Spark running on 32 machines. The logs include normal and anomalous executions, but are not labeled. \textbf{ZooKeeper} is another Apache service used in distributed computing and configuration management. Loghub also comprises logs from conventional operating systems. The \textbf{Windows} log data set was collected on a laboratory Windows 7 machine by monitoring CBS (Component Based Servicing), which operates on a package and update level. Similarly, the \textbf{Android} data set was collected from a mobile phone in a laboratory setting. Both data sets are not labeled and do not involve any purposefully injected anomalies. Logs from the Linux operating system are provided in the disk images of the \textbf{Digital Corpora}, where failures such as invalid authentications occur in the data \cite{garfinkel2009bringing}.

When it comes to the detection of anomalous behavior, all aforementioned data sets mainly provide failure events that are generated as part of legitimate system usage. However, there are also data sets that instead involve events generated from malicious activities and thus enable evaluation of anomaly-based intrusion detection systems. For example, the \textbf{DFRWS 2009} data set contains system logs from Linux devices that involve data exfiltration, unauthorized accesses, as well as the use of backdoor software \cite{dfrws2009}. The \textbf{Honeynet 2010} \cite{honeynet2010} and \textbf{Honeynet 2011} \cite{honeynet2011} data sets comprise common log files from compromised Linux machines that were illegitimately accessed. The Public Security Log Sharing Site \cite{chuvakin} provides \textbf{Linux Security Logs} from diverse sources and affected by different real-world intrusions, such as brute-force attacks. Unfortunately, none of these security log files come with labels for malicious events and thus authors need to generate their own ground truths to evaluate their approaches on these data sets \cite{studiawan2021anomaly}. 

\subsubsection{Evaluation metrics} \label{metrics}

Quantitative evaluation (ER-2) of anomaly detection approaches typically revolves around counting the numbers of correctly detected anomalous samples as true positives ($TP$), incorrectly detected non-anomalous samples as false positives ($FP$), incorrectly undetected anomalous samples as false negatives ($FN$), and correctly undetected non-anomalous samples as true negatives ($TN$). In the most basic setting where events are labeled individually and samples represent single events (e.g., as in the BGL data set), it is relatively straightforward to evaluate detected events with binary classification \cite{farzad2019log, farzad2021two}. Some of the reviewed papers additionally consider a multi-class classification problem for data sets where different types of failures have distinct labels by computing the averages of evaluation metrics over all classes \cite{sun2020context} or plotting confusion matrices \cite{decker2020comparison}. 

Given that log events are sometimes aggregated with diverse methods prior to detection, it stands to reason that there are different ways to determine whether a sample is anomalous or not, and whether it counts as a correct detection or not. For example, aggregation of logs in windows could require to count detected events as true positives as long as they are close enough to the actual anomaly in the event sequence \cite{baril2020application, syngal2021server}. Since a majority of the reviewed papers rely on the HDFS data set where labels are only available for whole event sessions rather than single events, the most common method to compute aforementioned metrics relies on counting of in-/correctly identified non-/anomalous sessions \cite{chenunsupervised, du2021log, huang2020hitanomaly, li2020logspy, wang2021multi}.

Regardless of how the positive and negative samples are counted, almost all authors eventually evaluate their approaches using the well-known metrics precision ($P = \frac{TP}{TP + FP}$), recall or true positive rate ($R = TPR = \frac{TP}{TP + FN}$), false positive rate ($FPR = \frac{FP}{FP + TN}$) and F1-score ($F1 = \frac{2 \cdot P \cdot R}{P + R}$). Less common evaluation metrics are the accuracy ($ACC = \frac{TP + TN}{TP + TN + FP + FN}$) used in 15 publications as well as the area under curve which is computed for precision-recall-curves \cite{chen2020logtransfer} and receiver operator characteristic (ROC) curves \cite{wang2018anomaly, hirakawa2020software}. Other metrics that are more specific to deep learning applications are the number of model parameters \cite{guo2021anomaly, wang2022lightlog} and time to train models or run the detection (ER-3) \cite{liu2021lognads, gu2021research, qian2020anomaly, cheansunan2019detecting, decker2020comparison, xie2020attention}. Some authors also assess characteristics of their approaches that go beyond standard anomaly detection evaluations, for example, whether training on combinations of multiple data sets improves the overall performance of classification \cite{hashemi2021onelog} or whether their approaches are robust against changes of log patterns over time \cite{zhang2019robust, huang2020hitanomaly, hashemi2021onelog}.

\subsubsection{Benchmark approaches}

\begin{figure}
	\centering
	\includegraphics[width=1\columnwidth]{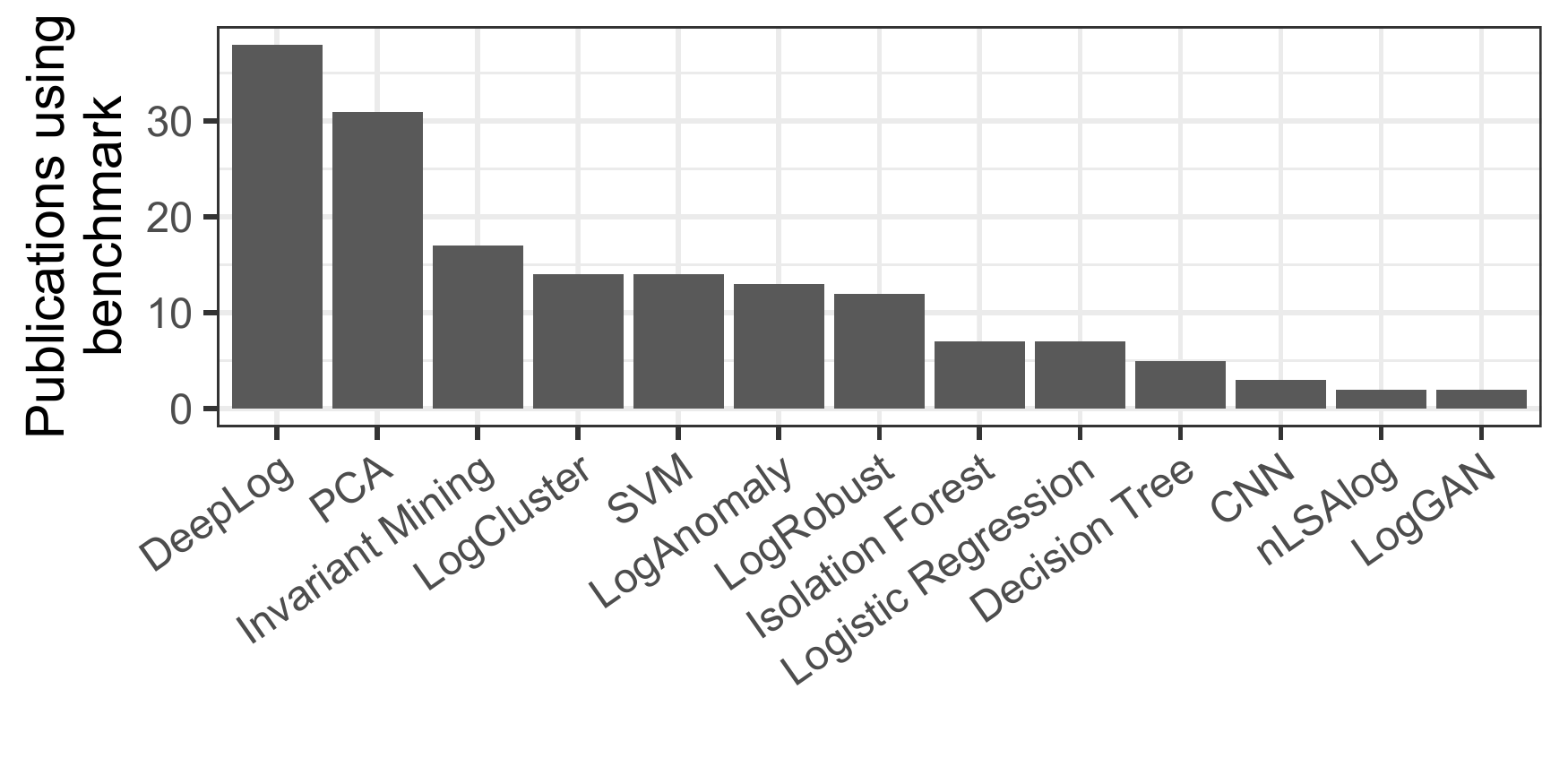}
	\caption{Overview of common benchmark approaches used in evaluations.}
	\label{fig:benchmarks}
\end{figure}

Most publications compare the evaluation metrics stated in the previous section with benchmark approaches to show their improvements over the state-of-the-art (ER-4). Figure \ref{fig:benchmarks} shows the most commonly used benchmarks in the reviewed publications. Note that we only considered approaches that appear in at least two different publications for this visualization. 

As visible in the figure, \textbf{DeepLog} \cite{du2017deeplog} is the most commonly used benchmark with appearances in 38 out of the 62 reviewed publications. DeepLog relies on LSTM RNNs to predict upcoming log events in sequences and thereby disclose observed events as anomalies if they are expected to occur with low probabilities. The popularity of DeepLog for comparisons can be explained by the facts that DeepLog was the first to detect sequential anomalies in log data using deep learning (cf. Sect. \ref{citations}) and that open-source re-implementations are available online. 

The second most commonly used benchmark leverages principal component analysis (\textbf{PCA}) and relies on event counts rather than sequences. In particular, Xu et al. \cite{xu2009largescale} create message count vectors for event type identifiers and use PCA to transform them into subspaces where anomalies appear as outliers that have a high distance to all other samples. Lou et al. \cite{lou2010mining} also generate message count vectors but use \textbf{Invariant Mining} to discover linear relationships between log events that represent execution workflows. Event sequences that violate previously identified invariants are declared as anomalies. 

\textbf{LogCluster} \cite{lin2016log} generates vectors for log sequences and then clusters them using a similarity metric. The approach is mainly designed for log filtering but can also be applied for detection of unusual log patterns. Support vector machines (\textbf{SVM}) \cite{liang2007failure} are yet another method relying on event count vectors. However, other than PCA and invariant mining, SVM typically operate in supervised manner. To alleviate this issue and enable application in semi- or unsupervised cases, authors therefore resort to one-class SVM \cite{scholkopf2001estimating} or Support Vector Data Description (SVDD) \cite{tax2004support}. 

Similar to DeepLog and contrary to aforementioned conventional machine learning approaches, \textbf{LogAnomaly} \cite{meng2019loganomaly} leverages LSTM RNNs to analyze log event sequences. To this end, they propose the so-called template2Vec embedding method to extract semantic vectors from the tokens that make up the events. \textbf{LogRobust} \cite{zhang2019robust} also makes use of semantic vectors but is specifically designed to handle unknown log events that appear as part of software evolution. 

\textbf{Isolation Forest} \cite{liu2008isolation} is an anomaly detection technique where the analyzed data is recursively split until a single data instance is isolated from all other points; thereby, anomalous points are identified as they are expected to require fewer splits until their isolation. \textbf{Logistic Regression} \cite{bodik2010fingerprinting} is a classifier for numeric input data that works best in linear classification cases \cite{studiawan2021anomaly}. \textbf{Decision Tree} \cite{chen2004failure} on the other hand is a supervised classification method where instances traverse a binary search tree. Thereby, each internal node splits the data by a specific predicate and each leaf of the tree determines the class of the instances. 

Another benchmark that relies on deep learning are \textbf{CNN}. In particular, the approach by Lu et al. \cite{lu2018detecting} semantically encodes sequences of event identifiers and embeds them into a matrix for convolution. Finally, \textbf{nLSAlog} \cite{yang2019nlsalog} leverages LSTM RNNs and is thus similar to DeepLog. Out of all reviewed publications, only eight do not involve any benchmark approach for comparison.

\subsubsection{Reproducibility} \label{reproduc}

Authors of scientific publications should pursue to enable reproducibility of their presented results for many reasons, including the possibility for others to validate the correctness of the approach, to extend the algorithms with additional features, to carry out their own experiments on other data sets, and to use the approach as benchmarks in new publications. We consider an approach reproducible when both the data used in the evaluation (ER-5) as well as the original source code (ER-6) are publicly available.

As outlined in Sect. \ref{datasets}, a majority of the reviewed publications carry out their evaluations on the same few data sets that are publicly available. Some authors also evaluate their approaches on private data sets that are synthetically generated in testbeds \cite{baril2020application, sundqvist2020boosted, zhang2021sentilog, yang2016log, wibisono2019log, li2020logspy, wang2021multi}, collected from academic institutions \cite{otomo2019latent, yang2021semi}, or obtained from industrial real-world applications \cite{zhang2019robust, wittkopp2021a2log, syngal2021server, yang2021semi, decker2020comparison, catillo2022autolog, chen2020logtransfer}. Overall, 55 out of the 62 reviewed publications involve evaluations on at least one of the publicly available data sets from Table \ref{tab:eval}.

While it is relatively common to evaluate approaches on public data sets, there are unfortunately only few authors that publish implementations of their approaches alongside the papers. During our review we were only able to find the original source code of presented approaches from 8 publications. However, we also point out that there exist some re-implementations of scientific approaches in the deep-loglizer toolbox provided by Chen et al. \cite{chen2021experience}. We mark approaches where implementations by the original authors exist with (YES), re-implementations by other authors as (RE), and all others as (NO) in Table \ref{results}. We encourage authors to publish their code to improve the reproducibility of their results and hope to see more open-source implementations in the future.

\section{Discussion} \label{discussion}

The previous sections presented the results of our survey in detail. In the following we summarize these results, discuss open issues in the research area on log-based anomaly detection using deep learning, and propose ideas for future research in course of answering our research questions from Sect. \ref{intro}.

\textit{\textbf{RQ1}: What are the main challenges of log-based anomaly detection with deep learning?} When carrying out our systematic literature review we assessed whether and to what degree the current state-of-the-art addresses the research challenges enumerated in Sect. \ref{challenges}. It turns out that data instability, i.e., the appearance of previously unknown events, is one of the main issues addressed by the reviewed approaches. The key idea to resolving this problem is currently to represent logs as semantic vectors so that new or changed events can still be compared to known events by measuring their similarities \cite{le2021log, zhang2019robust, lv2021conanomaly, ott2021robust, li2020swisslog, syngal2021server, yang2021semi, xi2021anomaly, han2021unsupervised, zhang2021logattn, catillo2022autolog}. There are many techniques for generating numeric vectors to represent log events (cf. Sect. \ref{featurerepr}) and thus resolve the issue of feeding unstructured and textual input data to neural networks. Imbalanced data sets are another challenge that is specifically addressed by some approaches. In particular, authors suggest to use sampling techniques as well as context-aware embedding methods as possible solutions \cite{wang2021multi, farzad2020log, xia2021loggan, sun2020context, li2020logspy}.

Some approaches are specifically designed to enable applicability in scenarios that demand efficient and lightweight algorithms, e.g., deployment on edge devices. This is achieved by leveraging low-dimensional vector representations as well as convolutional neural networks that are more efficient than recurrent neural networks \cite{wang2022lightlog, guo2021anomaly, cheansunan2019detecting}. Similarly, some approaches support log stream processing and enable adaptive learning (i.e., dynamically changing the baselines for anomaly detection) by incrementally re-training the models with manually identified and labeled false positive samples \cite{yen2019causalconvlstm}. It must be noted that there is currently no solution how to automatically determine that re-training is required without label information or manual intervention. The challenge of interleaving logs is generally solved by leveraging session identifiers directly from parsed log data; however, there are also approaches that evade the need for sequences altogether, e.g., by relying on sentiment analysis \cite{zhang2021sentilog}. 

A way to address the challenge that only few labeled data is available is provided by transfer learning, where models are trained on one system and tested on another \cite{guo2021translog, chen2020logtransfer}. The main idea is that the log event patterns learned by the neural networks are similar across different domains and that already seen anomalies can be recognized and classified. Guo et al. \cite{guo2021anomaly} are the only authors to consider federated learning, where learning takes place in a distributed manner across multiple systems. Hashemi et al. \cite{hashemi2021onelog} also go into this direction as they combine multiple data sets to evaluate whether this affects the performance of their model. We believe that federated learning could be an interesting topic for future publications as there exist many real-world scenarios where log data is monitored in distributed machines but orchestration of deployed detectors takes place centrally \cite{preuveneers2018chained}.

The challenge of facing diverse artifacts of anomalies is only partially addressed since the vast majority of approaches focus on sequences and frequencies of log events, but only few consider event parameters or inter-arrival times for detection. We recommend to also consider techniques that address other patterns that appear in normal system behavior and may be useful to detect specific anomalies, such as changes of parameter value correlations, periodic behavior, statistical distributions, etc. Moreover, we observed that the explainability of the proposed deep learning models is relatively low, i.e., it is non-trivial to understand the criteria for classifications and thus detection of model bias as well as interpretation of false positives and false negatives is generally difficult. This hinders root cause analysis of detected anomalies and produces an overhead for system operators. Specifically in security-critical systems (e.g., intrusion detection) it is vital to understand the functioning - and thus the limits - of deployed anomaly detectors. We would therefore recommend to direct future research in log-based anomaly detection towards explainable artificial intelligence \cite{arrieta2020explainable}.

\textit{\textbf{RQ2}: What state-of-the-art deep learning algorithms are typically applied?} Our review shows that diverse types of deep learning algorithms are used in scientific publications and that it is common to combine several approaches. Thereby, RNNs are clearly the most applied models, because they are a natural choice for capturing sequential patterns in log data. CNNs are used as an efficient alternative to RNNs as they are also able to pick up event dependencies. On the other hand, Autoencoders and Transformers are frequently applied as they support unsupervised learning. While GANs, MLPs, GNNs, and EGNNs are only used by few approaches, they have beneficial properties (cf. Sect. \ref{models}) that make these models worth considering. Similarly, we are convinced that other deep learning architectures that are not explored in the reviewed literature could yield interesting insights, e.g., deep belief networks or deep reinforcement learning \cite{sarker2021deepcyber, sarker2021deeplearning}.

We believe that the application of specific techniques is mostly motivated by the log data to be analyzed and anomalies to be detected. The fact that all of the commonly used log data sets involve anomalies that manifest as sequentially occurring events (cf. Sect. \ref{datasets}) thus explains the tendency towards RNNs. However, as anomalies could also manifest as point or contextual anomalies (cf. Sect. \ref{anomalydetection}) we recommend to consider alternative log data sets with different types of anomalies and to develop approaches for these cases. For example, in our earlier works \cite{landauer2020have, landauer2022maintainable} we published log data sets where anomalies affect combinations, compositions, and distributions of event parameter values in addition to frequencies and sequences of log events.

\textit{\textbf{RQ3}: How is log data pre-processed to be ingested by deep learning models?} Our survey shows that there are three main ways to approach feature extraction from raw log data. First, by tokenizing log messages, which is a simple method that does not require any parsers but lacks semantic interpretation of the tokens. Second, by parsing the messages and extracting information from collections of log events, such as sequences, counts, or statistics. Third, by extracting parameters including the time stamps from parsed log events. There are a multitude of methods to represent these features as numeric vectors to be ingested by deep learning models.

While traditional machine learning methods such as SVM or PCA work best with event count vectors, most approaches leveraging deep learning neural networks use semantic vectors to yield the best results \cite{wang2021log}. Thereby, the tokens that make up the log messages are represented as numeric vectors and considered in the context of their sequence of event occurrences. Most approaches employ these sequential features, while frequencies, one-hot encoded data, and embedding layers are used less often or only as a contributing feature.

\textit{\textbf{RQ4}: What types of anomalies are detected and how are they identified as such?} Almost all reviewed approaches focus on sequential anomalies that either manifest in the sequences of events, the sequences of tokens within events, or a combination of both. Only few approaches make use of event counts or detect single log lines as outliers without their context of occurrence. The detection technique is generally driven by the output of the neural networks. While binary or multi-class classifications are directly used to report anomalies, all numeric outputs such as anomaly scores or reconstruction errors are compared against pre-defined thresholds and probability distributions of log events are used to check whether the actual events is within the top candidates. Determining these thresholds is usually carried out empirically for a particular log file.

\textit{\textbf{RQ5}: How are the proposed models evaluated?} Our survey on commonly used log data sets from Sect. \ref{datasets} shows that there are only four data sets that are used by a majority of the approaches: HDFS, BGL, Thunderbird, and OpenStack. All these data sets are collected from scenarios involving high-performance computing and virtual machines that are affected by randomly occurring failures. It is obvious that the availability of anomaly labels make these data sets particularly attractive for scientific evaluations. As mentioned before, we argue that a larger and more diverse set of input data sets would be beneficial to evaluate whether the proposed approaches are capable of detecting anomalous artifacts other than unusual sequences. In particular, the consequences of cyber attacks rather than failures could result in log artifacts that are suitable for detection and an appropriate use-case for anomaly detection.

We manually analyzed the HDFS data set as it is the most popular of all and found that it is far from challenging to achieve competitive detection rates. The reason for this is that many of the anomalous event sequences are trivial to identify as they involve event types that never occur in the training data or involve fewer elements than the shortest normal sequences. Using these two heuristics we are able to achieve $F1=90.41$\%, $ACC=99.48$\%, $P=98.37$\%, $R=83.65$\%, $FPR=0.04$\% on the test data. Moreover, using these heuristics in combination with the simple Stide algorithm \cite{forrest1996sense} that moves a sliding window of a given length over the data and looks for sub-sequences that have not been seen before in the training data further improves the evaluation metrics to $F1=95.14$\%, $ACC=99.71$\%, $P=95.27$\%, $R=95.01$\%, $FPR=0.14$\% when using a window size of $2$. Omitting sequential information altogether and only leveraging similarities of event count vectors further pushes these evaluation metrics to $F1=98.86$\%, $ACC=99.93$\%, $P=97.81$\%, $R=99.92$\%, $FPR=00.07$\%. For comparison, DeepLog only yields detection scores of $F1=95.72$\%, $ACC=99.75$\%, $P=95.12$\%, $R=96.32$\%, $FPR=0.15$\% \cite{du2017deeplog}. We provide the code for our experiments in a reproducible form as open-source implementations (in separate repositories for Stide\footnote{\url{https://github.com/ait-aecid/stide}} and similarity-based event count vector clustering\footnote{\url{https://github.com/ait-aecid/count-vector-clustering}}). It is not clear to us why such approaches were not used as benchmarks in any of the reviewed publications: They only take a fraction of the time for training and processing the test data in comparison to deep (and also most conventional) learning models, and additionally have a much better explainability than neural networks. Similar conclusions have been drawn for the case of log event prediction using the HDFS data set \cite{mantyla2022pinpointing}. We argue that the fact that such simple algorithms achieve competitive detection rates to deep learning models further urges authors to consider additional data sets where more diverse anomalous artifacts are present and the benefits of their approaches such as robustness against data instability become apparent.

Another issue that we noticed is that evaluations are usually carried out on the basis of anomalous sequences, i.e., the whole sequence is considered normal or anomalous rather than its elements \cite{du2017deeplog}. However, parts of long sequences may actually represent normal system behavior while only a few elements should be considered anomalies. It would be interesting to evaluate whether detection approaches are able to pinpoint exactly which parts of sequences are anomalous, which would also be practical for manual investigations of reported anomalies by system operators. Obviously this requires that data sets are labeled on the granularity of single events rather than sessions \cite{mantyla2022pinpointing}.

Finally, we note that almost all evaluations rely on metrics such as the F-score (cf. Sect. \ref{metrics}) that are known to not accurately depict the classification or detection performance when data sets are highly imbalanced. To avoid misinterpretations of evaluation results, it is recommended to also compute metrics that are more robust against class imbalance, such as the specificity or true negative rate \cite{le2022log}. 

\textit{\textbf{RQ6}: To what extent do the approaches rely on labeled data and support incremental learning?} Log-based anomaly detection is most often applied in use-cases that aim to disclose unexpected system behavior such as failures or cyber attacks. Since these artifacts are not known beforehand and thus no labels can exist, un- or semi-supervised approaches are generally more widely applicable and therefore preferable \cite{chandola2009anomaly}. Since semi-supervised learning can be achieved with most neural network architectures including RNNs, but only specific deep learning models support fully unsupervised operation, we find 28 semi-supervised approaches in our reviewed literature as opposed to only 8 out of 62 approaches that are unsupervised. We did not expect to see the relatively large amount of 26 supervised approaches that require at least partially labeled anomalies for training. Moreover, with 54 out of 62 a vast majority of approaches only support offline training. This includes most supervised models and also all other approaches that do not intend to dynamically and automatically update the trained models over time. Only 8 of the reviewed approaches enable continuous model adjustments through re-training or EGNN model architectures \cite{decker2020comparison}. 

\textit{\textbf{RQ7}: How reproducible are the presented results in terms of availability of source code and used data?} As pointed out in Sect. \ref{reproduc}, a majority of the reviewed approaches make use of publicly available data sets to evaluate their approaches and only 7 publications rely on private data sets. The reproducibility of the presented results is thus relatively high, assuming that readers are willing to re-implement the approaches based on the descriptions from the papers from scratch. We could only find 9 publicly available source codes of approaches published by the original authors as well as 4 re-implementations, indicating a low reproducibility overall. We encourage authors to publish reproducible experiments in the future to also enable large-scale quantitative comparisons in surveys.

\textbf{Recommendations}. Based on the aforementioned answers to our research questions and the identified issues, we propose the following research agenda: First of all, adequate and diverse log data sets with state-of-the-art benchmarks are needed to ensure applicability of deep learning in generic anomaly detection use-cases and demonstrate their superiority over simple or conventional detection methods. Second, low explainability of detection results is a primary concern that permeates the entire research field and needs to be addressed appropriately. Resolving these two issues largely improves the comprehensibility and reliability of proposed methods and facilitates the development of novel deep learning detection algorithms. With this research agenda in mind, we summarize our recommendations for future research as follows.
\begin{itemize}
	\item Create or identify new log data sets that specifically involve sequential anomalies and are less affected by other types of anomalies when evaluating approaches that ingest log data as sequences.
	\item Work on methods that improve the explainability of proposed approaches for anomaly detection using deep learning, for example, by the extraction of specific detection rules from the models or by determining the main features responsible for the detection of specific instances and augmenting detection results with that information.
	\item Consider log artifacts other than event sequences for anomaly detection or use them as additional input to deep learning models.
	\item Propose novel anomaly detection methods or deep learning architectures that resolve common challenges for practical applications, specifically regarding incremental and stream processing or log data, adaptive learning, as well as efficient and low-resource training.
	\item Consider pinpointing anomalies within sequences rather than detecting whole sequences as anomalous.
	\item Allow researchers to reproduce and extend presented results by publishing developed code as open-source and used log data sets on data sharing repositories.
\end{itemize}

\section{Conclusion} \label{conclusion}

This paper presents a survey of 62 scientific approaches that pursue the detection of anomalous events or processes in system log data using deep learning. The survey shows that diverse model architectures are suitable for this purpose, including models for sequential input data such as recurrent or convolutional neural networks, language-based models such as transformers, as well as unsupervised models such as Autoencoders or generative adversarial networks. Similarly, there are different features used for training and subsequently for detection, such as sequences and counts of events or tokens as well as parameter values or statistics derived from the events. To enable processing of these features as input to neural networks it is necessary to encode them as numeric vectors, for example, through semantic vectorization or one-hot encoding. Anomalies are then detected either directly through classification or by deriving some kind of anomaly score from the network that allows to discern normal from anomalous system behavior. The survey shows that there are open challenges that are not sufficiently resolved by existing approaches, including detection techniques that go beyond sequential anomalies, low explainability of trained models and classification results, lack of representative evaluation data sets containing diverse attack artifacts, and a low reproducibility.

\section*{Acknowledgments}
This work was partly funded by the European Defence Fund (EDF) projects AInception (101103385) and PANDORA (SI2.835928), and the FFG project DECEPT (873980).

\bibliographystyle{IEEEtran}
\bibliography{refs} 

\end{document}